\RequirePackage{fix-cm}
\documentclass[smallcondensed]{svjour3}     
\smartqed  
\usepackage{booktabs}
\usepackage{multicol}
\usepackage{enumitem}
\usepackage{longtable}
\usepackage{todonotes}
\usepackage{amsmath}
\usepackage{graphicx}
\usepackage[utf8]{inputenc}
\usepackage{tikz}
\usepackage{mwe}
\usepackage{subcaption}
\usepackage{rotating}
\usepackage{pgfplots}
\pgfplotsset{width=7cm, compat=1.10}
\usepgfplotslibrary{fillbetween}
\usepackage{afterpage}
\usepackage{pdflscape}
\renewcommand{\thefigure}{\arabic{figure}}
\newcommand{\pkg}[1]{{\fontseries{b}\selectfont #1}} 
\usepackage{bbm}
\usepackage{algorithm}
\usepackage{natbib}
\usepackage[noend]{algpseudocode}
\usepackage[bottom=1.2in, top =1.2in, left=1in, right=1in]{geometry}

\makeatletter
\def\BState{\State\hskip-\ALG@thistlm}
\makeatother

\tikzset{
  treenode/.style = {shape=rectangle, rounded corners,
                     draw, align=center,
                     fill=blue!10},
  root/.style     = {treenode, font=\small, fill=blue!5},
  leaf/.style      = {treenode, font=\small},
  dummy/.style    = {shape=rectangle,draw=white,align=left}
}

\journalname{Machine Learning}
\begin{document}

\title{Optimal Survival Trees
}

\author{Dimitris~Bertsimas \and Jack~Dunn \and Emma~Gibson \and Agni~Orfanoudaki }

\institute{Dimitris Bertsimas \at
          Operations Research Center, E40-111,
          Massachusetts Institute of Technology,
          Cambridge, MA 02139, USA.
          \email{dbertim@mit.edu}
          \and
          Jack Dunn \at 
          Operations Research Center, E40-111, 
          Massachusetts Institute of Technology
          Cambridge, MA 02139, USA.
          \email{jack@interpretable.ai} 
          \and
           Emma Gibson \at 
          Operations Research Center, E40-111, 
          Massachusetts Institute of Technology
          Cambridge, MA 02139, USA.
          \email{emgibson@mit.edu} 
          \and
           Agni Orfanoudaki \at 
          Operations Research Center, E40-111, 
          Massachusetts Institute of Technology
          Cambridge, MA 02139, USA.
          \email{agniorf@mit.edu} 
}

\date{Received: date / Accepted: date}

\maketitle

\begin{abstract}
Tree-based models are increasingly popular due to their ability to identify complex relationships that are beyond the scope of parametric models. Survival tree methods adapt these models to allow for the analysis of censored outcomes, which often appear in medical data. We present a new Optimal Survival Trees algorithm that leverages mixed-integer optimization (MIO) and local search techniques to generate globally optimized survival tree models. We demonstrate that the OST algorithm improves on the accuracy of existing survival tree methods, particularly in large datasets.


\end{abstract}

\section{Introduction}\label{sec:intro}

Survival analysis is a cornerstone of healthcare research and is widely used in the analysis of clinical trials as well as large-scale medical datasets such as 
Electronic Health Records and insurance claims. Survival analysis methods are required for censored data in which the outcome of interest is generally the time until an event (onset of disease, death, etc.), but  the exact time of the event is unknown (censored) for some individuals. When a lower bound for these missing values is known (for example, a patient is known to be alive until at least time $t$) the data is said to be right-censored.

A common survival analysis technique is  Cox proportional hazards regression  \citep{cox1972regression} which models the hazard rate for an event as a linear combination of covariate effects. Although this model is widely used and easily interpreted, its parametric nature makes it unable to identify non-linear effects or interactions between covariates \citep{bou2011review}.

Recursive partitioning techniques (also referred to as \textit{trees}) are a popular alternative to parametric models. When applied to survival data, survival tree algorithms partition the covariate space into smaller and smaller regions (\textit{nodes}) containing observations with homogeneous survival outcomes. The survival distribution in the final partitions (\textit{leaves}) can be analyzed using a variety of statistical techniques such as Kaplan-Meier curve estimates \citep{kaplan1958nonparametric}. 

Most recursive partitioning algorithms generate trees in a top-down, greedy manner, which means that each split is selected in isolation without considering its effect on subsequent splits in the tree. However, Bertsimas and Dunn \citep{bertsimas2017optimal, bertsimas2017optimalBOOK} have proposed a new algorithm which uses modern mixed-integer optimization (MIO) techniques to form the entire decision tree in a single step, allowing each split to be determined with full knowledge of all other splits. This \textit{Optimal Trees} algorithm allows the construction of single decision trees for classification and regression that have performance comparable with state-of-the-art methods such as random forests and gradient boosted trees, without sacrificing the interpretability offered by a single-tree model. 
  
The key contributions of this paper are: 
\begin{enumerate}
  \item We present \textit{Optimal Survival Trees} (OST), a new survival trees algorithm that utilizes the \textit{Optimal Trees} framework to generate interpretable trees for censored data. 
  \item We propose a new accuracy metric that evaluates the fit of Kaplan-Meier curve estimates relative to known survival distributions in simulated datasets. We also demonstrate that this metric is consistent with the Integrated Brier Score \citep{graf1999assessment}, which can be used to evaluate the fit of Kaplan-Meier curves when the true distributions are unknown.
  \item We evaluate the performance of our method in both simulated and real-world datasets and demonstrate improved accuracy relative to two existing algorithms. 
  \item Finally, we provide an example of how the algorithm can be used to predict the risk of adverse events associated with cardiovascular health in 
the Framingham Heart Study (FHS) dataset. 
\end{enumerate}

The structure of this paper is as follows. We review existing survival tree algorithms in Section~\ref{sec:survival_lit} and discuss some of the technical challenges associated with building trees for censored data. In Section~\ref{sec:predictivetrees}, we give an overview of the Optimal Trees algorithm proposed by Bertsimas and Dunn and we adapt this algorithm for Optimal Survival Trees in Section~\ref{sec:survivaltrees}. Section~\ref{sec:metrics} begins with a discussion of existing survival tree accuracy metrics, followed by the new accuracy metrics that we have introduced to evaluate survival tree models in simulated datasets.  Simulation results are presented in Section~\ref{sec:simulations} and results for real-world datasets are presented in Sections~\ref{sec:realdatasets}--\ref{sec:fhs}. We conclude in Section~\ref{sec:conclusion} with a brief summary of our contributions.

\section{Review of Survival Trees} \label{sec:survival_lit}

Recursive partitioning methods have received a great deal of attention in the literature, the most prominent method being the Classification and Regression Tree algorithm (CART) \citep{breiman1984classification}. Tree-based models are appealing due to their logical, interpretable structure as well as their ability to detect complex interactions between covariates. However, traditional tree algorithms require complete observations of the dependent variable in training data, making them unsuitable for censored data.

Tree algorithms incorporate a splitting rule which selects partitions to add to the tree, and a pruning rule determines when to stop adding further partitions. 
Since the 1980s, many authors have proposed splitting and pruning rules for censored data. Splitting rules in survival trees are generally based on either (a) node distance measures that seek to maximize the difference between observations in separate nodes  or (b) node purity measures that seek to group similar observation in a single node \citep{zhou2015rationale, molinaro2004tree}.

Algorithms based on node distance measures  compare the two adjacent child nodes that are generated when a parent node is split, retaining the split that produces the greatest difference in the child nodes. Proposed measures of node distance include the  two-sample logrank test \citep{ciampi1986stratification},  the likelihood ratio statistic \citep{ciampi1987recursive} and conditional inference permutation tests \citep{hothorn2006unbiased}. We note that the score function used in Cox regression models also falls into the class of node distance measures, as the partial likelihood statistic is based on a comparison of the relative risk coefficient predicted for each observation.

Dissimilarity-based splitting rules are unsuitable for certain applications (such as the Optimal Trees algorithm) because they do not allow for the assessment of a single node in isolation. We will therefore focus on node purity splitting rules for developing the OST algorithm.

\citet{gordon1985tree} published the first survival tree algorithm with a node purity splitting rule based on Kaplan-Meier estimates. \citet{davis1989exponential} used a splitting rule based on the negative log-likelihood of an exponential model, while
\citet{therneau1990martingale} proposed using martingale residuals   as an estimate of node error. \citet{leblanc1992relative} suggested comparing the log-likelihood of a saturated model to the first step of a full likelihood estimation procedure for the proportional hazards model and showed that both the full likelihood and martingale residuals can be calculated efficiently from the Nelson-Aalen cumulative hazard estimator \citep{nelson1972theory, aalen1978nonparametric}. More recently, \citet{molinaro2004tree} proposed a new approach to adjust loss functions for uncensored data based on inverse probability of censoring weights (IPCW).

Most survival tree algorithms make use of cost-complexity pruning to determine the correct tree size, particularly when node purity splitting is used. Cost-complexity pruning selects a tree that minimizes a weighted combination of the total tree error (i.e., the sum of each leaf node error) and tree complexity (the number of leaf nodes), with relative weights determined by cross-validation. 
A similar split-complexity pruning method was suggested by \citet{leblanc1993survival} for node distance measures, using the sum of the split test statistics and the number of splits in the tree. Other proposals include using the Akaike Information Criterion (AIC) \citep{ciampi1986stratification} or using a $p$-value stopping criterion to stop growing the tree when no further significant splits are found \citep{hothorn2006unbiased}.

Survival tree methods have been extended to include ``survival forest'' algorithms which  aggregate the results of multiple trees. \citet{breiman2002} adapted the CART-based random forest algorithm to survival data, while both \citet{hothorn2004bagging} and \citet{ishwaran2008random} proposed more general methods that generate survival forests from any survival tree algorithm. The aim of survival forest models is to produce more accurate predictions by avoiding the instability of single-tree models. However, this approach leads to ``black-box'' models which are not interpretable and therefore lack one of the primary advantages of single-tree models.

Relatively few survival tree algorithms have been implemented in publicly available, well-documented software. Two user-friendly options are available in \pkg{R} \citep{R2017} packages: Therneau's algorithm based on martingale residuals  is implemented in the \pkg{rpart} package \citep{therneau2010rpart}    and Hothorn's conditional inference (\pkg{ctree}) algorithm in the \pkg{party}  package \citep{hothorn2010party}.

\section{Review of Optimal Predictive Trees}\label{sec:predictivetrees}

In this section, we briefly review approaches to constructing decision trees, and in particular, we outline the Optimal Trees algorithm. The purpose of this section is to provide a high-level overview of the Optimal Trees framework; interested readers are encouraged to refer to~\citet{bertsimas2017optimalBOOK} and \citet{dunnthesis} for more detailed technical information. The Optimal Trees algorithm and is implemented in \texttt{Julia} \citep{bezanson2017julia} and is available to academic researchers under a free academic license.\footnote{Please email survival-trees@mit.edu to request an academic license for the Optimal Survival Trees package.}

Traditionally, decision trees are trained using a greedy heuristic that recursively partitions the feature space using a sequence of locally-optimal splits to construct a tree. This approach is used by methods like CART~\citep{breiman1984classification} to find classification and regression trees. The greediness of this approach is also its main drawback---each split in the tree is determined independently without considering the possible impact of future splits in the tree on the quality of the here-and-now decision. This can create difficulties in learning the true underlying patterns in the data and lead to trees that generalize poorly. The most natural way to address this limitation is to consider forming the decision tree in a single step, where each split in the tree is decided with full knowledge of all other splits.

Optimal Trees is a novel approach for decision tree construction that significantly outperforms existing decision tree methods \citep{bertsimas2017optimalBOOK}. It formulates the decision tree construction problem from the perspective of global optimality using mixed-integer optimization (MIO), and solves this problem with coordinate descent to find optimal or near-optimal solutions in practical run times. These Optimal Trees are often as powerful as state-of-the-art methods like random forests or boosted trees, yet they are just a single decision tree and hence are readily interpretable. This obviates the need to trade off between interpretability and state-of-the-art accuracy when choosing a predictive method.

The Optimal Trees framework is a generic approach that tractably and efficiently trains decision trees according to a loss function of the form
\begin{equation}\label{eq:generic_obj}
	\min_T ~~\texttt{error}(T, D) + \alpha \cdot \texttt{complexity}(T),
\end{equation}
where $T$ is the decision tree being optimized, $D$ is the training data, $\texttt{error}(T, D)$ is a function measuring how well the tree $T$ fits the training data $D$, 
\texttt{complexity}$(T)$ is a function penalizing the complexity of the tree (for a tree with splits parallel to the axis, this is simply the number of splits in the tree), and $\alpha$ is the \emph{complexity parameter} that controls the tradeoff between the quality of the fit and the size of the tree.

There have been many attempts in the literature to construct globally optimal predictive trees \citep{bennett1996optimal,son1998optimal,grubinger2014evtree}. However, these methods could not scale to datasets of the sizes required by practical applications, and therefore did not displace greedy heuristics as the approach used in practice. Unlike the others, Optimal Trees is able scale to large datasets ($n$ in the millions, $p$ in the thousands) by using coordinate descent to train the decision trees towards global optimality. When training a tree, the splits in the tree are repeatedly optimized one-at-a-time, finding changes that improve the global objective value in Problem~\eqref{eq:generic_obj}. To give a high-level overview, the nodes of the tree are visited in a random order and at each node we consider the following modifications:
\begin{itemize}
  \item If the node is not a leaf, delete the split at that node;
  \item If the node is not a leaf, find the optimal split to use at that node and update the current split;
  \item If the node is a leaf, create a new split at that node.
\end{itemize}

For each of the changes, we calculate the objective value of the modified tree with respect to Problem~\eqref{eq:generic_obj}. If any of these changes result in an improved objective value, then the modification is accepted. When a modification is accepted or all potential modifications have been dismissed, the algorithm proceeds to visit the nodes of the tree in a random order until no further improvements are found, meaning that this tree is a locally optimal for Problem~\eqref{eq:generic_obj}. The problem is non-convex, so we repeat the coordinate descent process from various randomly-generated starting decision trees, before selecting the final locally-optimal tree with the lowest overall objective value as the best solution. For a more comprehensive guide to the coordinate descent process, we refer the reader to \citet{bertsimas2017optimalBOOK}.

Although only one tree model is ultimately selected, information from multiple trees generated during the training process is also used to improve the performance of the algorithm. For example, the Optimal Trees algorithm combines the result of multiple trees to automatically calibrate the complexity parameter ($\alpha$)  and to calculate variable importance scores in the same way as random forests or boosted trees. More detailed explanations of these procedures can be found in \citet{dunnthesis}.

\begin{figure}
  \centering
  \includegraphics[width=0.8\textwidth]{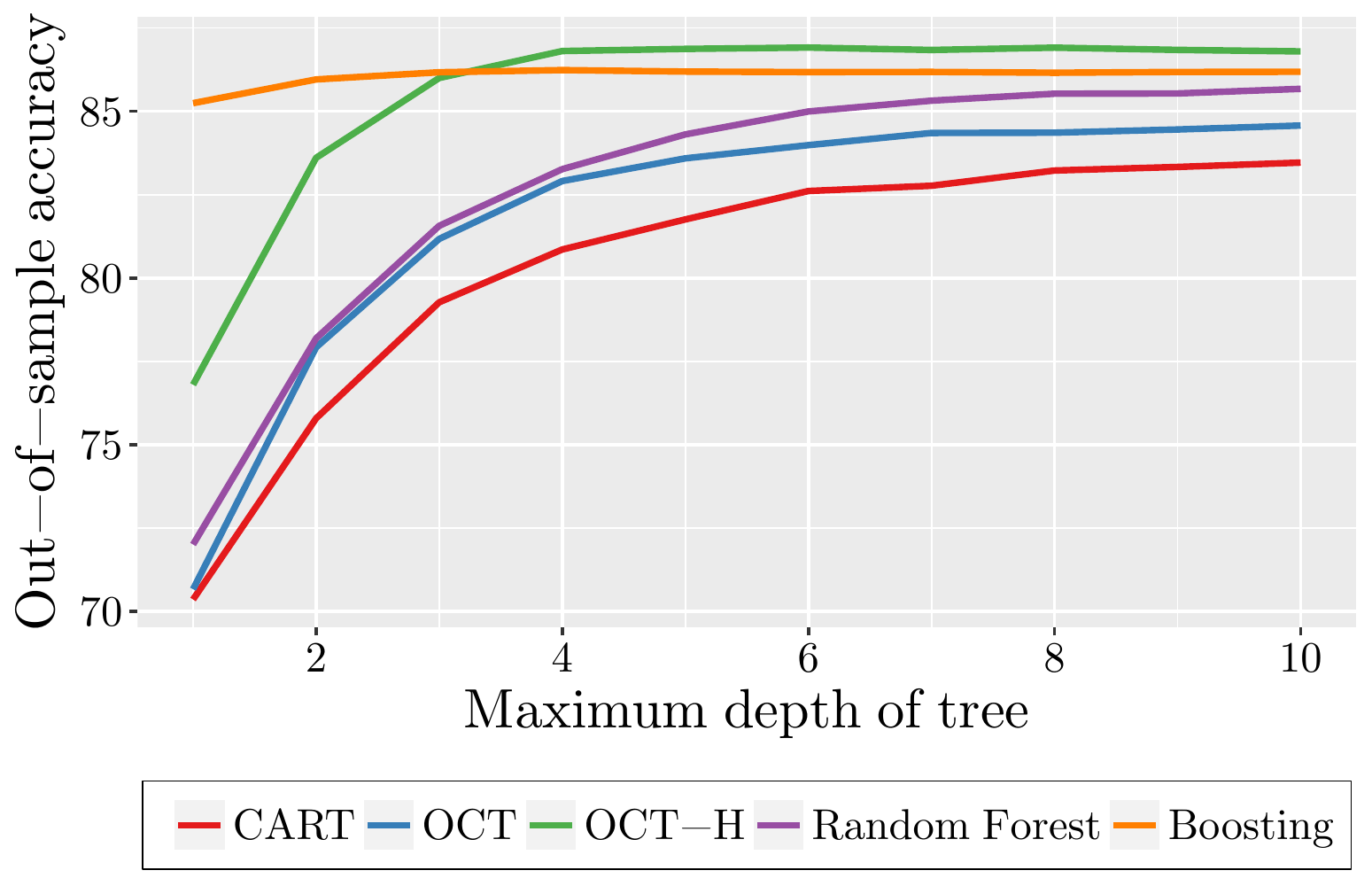}
  \caption{Performance of classification methods averaged across 60 real-world datasets. OCT and OCT-H refer to Optimal Classification Trees without and with hyperplane splits, respectively.}
  \label{fig:prediction-results}
\end{figure}

The coordinate descent approach used by Optimal Trees is generic and can be applied to optimize a decision tree under any objective function. For example, the Optimal Trees framework can train Optimal Classification Trees (OCT) by setting $\texttt{error}(T, D)$ to be the misclassification error associated with the tree predictions made on the training data. We provide a comparison of performance between various classification methods from~\citet{bertsimas2017optimalBOOK} in Figure~\ref{fig:prediction-results}. This comparison shows the performance of two versions of Optimal Classification Trees: OCT with parallel splits (using one variable in each split); and OCT with hyperplane splits (using a linear combination of variables in each split). These results demonstrate that not only do the Optimal Tree methods significantly outperform CART in producing a single predictive tree, but also that these trees have performance comparable with some of the best classification methods.

In Section~\ref{sec:survivaltrees}, we will extend the Optimal Trees framework to work with censored data and generate Optimal Survival Trees.

\section{Survival tree algorithm}\label{sec:survivaltrees}

In this section, we adapt the Optimal Trees algorithm described in Section~\ref{sec:predictivetrees} for the analysis of censored data. For simplicity, we will use terminology from survival analysis and assume that the outcome of interest is the time until death.  We begin with a set of observations $(t_i,\delta_i)_{i=1}^n$ where $t_i$ indicates the time of last observation and $\delta_i$ indicates whether the observation was a death ($\delta_i=1$) or a censoring ($\delta_i=0$).

Like other tree algorithms, the OST model requires a target function that determines which splits should be added to the tree. Computational efficiency is an important factor in the choice of target function, since it must be re-evaluated for every potential change to the tree during the optimization procedures. A key requirement for the target function is that the ``fit'' or error of each node should be evaluated independently of the rest of the tree. In this case, changing a particular split in the tree will only require re-evaluation of the subtree directly below that split, rather than the entire tree. This requirement restricts the choice of target function to the node purity approaches described in Section~\ref{sec:survival_lit}.

The splitting rule implemented in the OST algorithm is based on the likelihood method proposed by \citet{leblanc1992relative}. This splitting rule is derived from a proportional hazards model which assumes that the underlying survival distribution for each observation is given by %
\begin{equation}\label{eqn:CDF}
\mathrm{P}(S_i\leq t) = 1-e^{-\theta_i\Lambda(t)},
\end{equation}  
where $\Lambda(t)$ is the baseline cumulative hazard function and the coefficients $\theta_i$ are the  adjustments to the baseline cumulative hazard for each observation. 

In a survival tree model we replace $\Lambda(t)$ with an empirical estimate for the cumulative probability of death at each of the observation times. This is known as the Nelson-Aalen estimator \citep{nelson1972theory, aalen1978nonparametric},
\begin{equation} \label{eqn:Nels}\hat{\Lambda}(t) = \sum_{i:t_i\leq t}\frac{\delta_i}{\sum_{j:t_j\geq t_i} 1}.\end{equation}
Assuming this baseline hazard, the objective of the survival tree model is to optimize the hazard coefficients $\theta_i$. We impose that the tree model uses the same coefficient for all observations contained in a given leaf node in the tree, i.e. $\theta_i = \hat{\theta}_{T(i)}$. These coefficients are determined by maximizing the within-leaf sample likelihood 
\begin{equation}\label{eqn:Likelihood}
L= \prod\limits_{i=1}^n \left(\theta_i\frac{d}{dt}\Lambda(t_i)\right)^{\delta_i}e^{-\theta_i\Lambda(t_i)},
\end{equation}  
to obtain the node coefficients \begin{equation}\label{eqn:MLE} 
\hat{\theta}_{k} =\frac{\sum_{i}\delta_i I_{\{T_i = k\}}}{ \sum_{i}\hat{\Lambda}(t_i)I_{\{T_i = k\}}}.
\end{equation}
To evaluate how well different splits fit the available data we compare the current tree model to a tree with a single coefficient for each observation. We will refer to this as a fully saturated tree, since it has a unique parameter for every observation. The maximum likelihood estimates for these saturated model coefficients are \begin{equation} \label{eqn:sat}\hat{\theta}^{sat}_i = \frac{\delta_i}{\hat{\Lambda}(t_i)},\quad i=1,\dots, n.\end{equation}
We calculate the prediction error at each node as the difference between the log-likelihood for the fitted node coefficient and the saturated model coefficients at that node:
\begin{equation}\label{eqn:error}
\texttt{error}_k =\sum_{i:T(i) = k} \left(\delta_i \log\left(\dfrac{\delta_i}{ \hat{\Lambda}(t_i)}\right) - \delta_i \log(\hat{\theta}_{k})- \delta_i +\hat{\Lambda}(t_i)\hat{\theta}_{k}\right).
\end{equation}

The overall error function used to optimize the tree is simply the sum of the errors across the leaf nodes of the tree $T$ given the training data $D$:
\begin{equation}\label{eqn:error_total}
\texttt{error}(T, D) = \sum_{k \in \mathrm{leaves}(T)} \texttt{error}_k(D).
\end{equation}

We can then apply the Optimal Trees approach to train a tree according to this error function by substituting this expression into the overall loss function~\eqref{eq:generic_obj}. At each step of the coordinate descent process, we determine new estimates for $\hat{\theta}_{k}$ for each leaf node $k$ in the tree using~\eqref{eqn:MLE}. We then calculate and sum the errors at each node using~\eqref{eqn:error} to obtain the total error of the current solution, which is used to guide the coordinate descent and generate trees that minimize the error~\eqref{eqn:error_total}.

\section{Survival tree accuracy metrics} \label{sec:metrics}
In order to assess the performance of the OST algorithm, we now introduce a number of accuracy metrics for survival tree models. 
We will use the notation $T$ to represent a tree model, where $T_i = T(X_i)$ is the leaf node classification of observation $i$ with covariates $X_i$ in the tree T. We will use the notation $T^0$ to represent a null model (a tree with no splits and a single node).

\subsection{Review of survival model metrics}\label{sec:reviewmetrics}

We begin by reviewing existing accuracy metrics for survival models that are commonly used in both the literature as well as practical applications.

\begin{enumerate}
\item \textbf{Cox Partial Likelihood Score}

The Cox proportional hazards model \citep{cox1972regression} is a semi-parametric model that is widely used in survival analysis. The Cox hazard function estimate is
\begin{equation} \lambda(t|X_i) = \lambda_0(t)\exp{(\beta_1X_{i1} + \dots +\beta_pX_{ip})} = \lambda_0(t)\exp{(\beta^TX_i)} ,\end{equation} 
where $\lambda_0(t)$ is the baseline hazard function and $\beta$ is a vector of fitted coefficients.
This proportional hazards model does not make any assumptions about the form of $\lambda_0(t)$, and its parameters can be estimated  even when the baseline is completely unknown \citep{cox1975partial}. The coefficients $\beta$ are estimated by maximizing the partial likelihood function\footnote{This definition of the partial likelihood assumes that there are no ties in the data set (i.e., no two subjects have the same event time).},
\begin{equation} L(\beta) = \prod_{\mathrm{t_i uncensored}} \frac{\exp{(X_i\beta)}}{\sum_{t_j\geq t_i}\exp{(X_j\beta)}}= 
\prod_{\mathrm{t_i uncensored}}\frac{\theta_i}{\sum_{t_j\geq t_i} \theta_j}.\end{equation} 

For computational convenience, the Cox model is generally implemented using the log partial likelihood,
\begin{equation} l(\beta) = \log L(\beta)  =\sum_{\mathrm{t_i uncensored}} X_i\beta - \log(\sum_{t_j\geq t_i}\exp{(X_j\beta)}).\end{equation} 


In the context of survival trees, we can find the Cox hazard function associated with a particular tree model by assigning one coefficient to each leaf node in the tree, i.e.,
\begin{equation} \lambda_T(t) = \lambda_0(t)\exp{(\sum_{k \in T}\beta_k\mathbbm{1}(T_i=k)) } = \lambda_0(t)\exp{(\beta_{T_i})} .\end{equation}

We define the Cox Score for a tree model as the maximized log partial likelihood for the associated Cox model, $\max_{\beta}l(\beta|T)$. To assist with interpretation, we also define the Cox Score Ratio (CSR) as the percentage reduction in the Cox Score for tree $T$ relative to a null model,
\begin{equation} CSR(T) = 1-\frac{\max_{\beta}l(\beta|T)}{\max_{\beta}l(\beta|T^0)} .\end{equation} 

\item \textbf{The Concordance Statistic}

Applying a ranking approach to survival analysis is an effective way to deal with the skewed distributions of survival times as well as censored of the data. The Concordance Statistic, which is most familiar from logistic regression, is another popular metric that has been adapted to measure goodness-of-fit in survival models \citep{harrell1982evaluating}. The concordance index is defined as the proportion of all \textit{comparable} pairs of observations in which the model's predictions are \textit{concordant} with the observed outcomes. 

Two observations are \textit{comparable} if it is know with certainty that one individual died before the other. This occurs when the actual time of death is observed for both individuals (neither is censored) or when the one individual's death is observed before the other is censored. A comparable pair is \textit{concordant} if the predicted risk is higher for the individual that died first, and the pair is discordant if the predicted risk is lower for the individual that died first.  
Thus, the number of concordant pairs in a sample is given by
\begin{equation} CC = \sum_{i,j} \mathbbm{1}(t_i > t_j)\mathbbm{1}(\theta_i < \theta_j)\delta_j,\end{equation} 
and the number of discordant pairs is
\begin{equation} DC = \sum_{i,j} \mathbbm{1}(t_i > t_j)\mathbbm{1}(\theta_i > \theta_j)\delta_j,\end{equation} 
where the indices $i$ and $j$ refer to pairs of observations in the sample. Multiplication by the factor $\delta_j$ discards pairs of observations that are not comparable because the smaller survival time is censored, i.e., $ \delta_j = 0$. These definitions do not include comparable pairs with tied risk predictions, so we denote these pairs as 
\begin{equation} TR = \sum_{i,j} \mathbbm{1}(t_i > t_j)\mathbbm{1}(\theta_i = \theta_j)\delta_j.\end{equation} 

The number of concordant and discordant pairs is commonly summarized using Harrell's C-index \citep{harrell1982evaluating},
\begin{equation} H_C = \frac{CC+0.5\times TR}{CC+DC+TR}.\end{equation} 
Harrell's C takes values between 0 and 1, with higher values indicating a better fit. Note that randomly assigned predictions have an expected score of $H_C=0.5$. 

More recently, \citet{uno2011c} introduced a  non-parametric C-Statistic,
\begin{equation} U_{C_\tau} = \frac{\sum_{i,j} (\hat{G}(t_j))^{-2}\mathbbm{1}(t_i > t_j,t_j<\tau)\mathbbm{1}(\theta_i < \theta_j)\delta_j}{\sum_{i,j}\hat((G)(t_j))^{-2} \mathbbm{1}(t_i > t_j,t_j<\tau)(\theta_i > \theta_j)(\delta_j)+(t_i > t_j,t_j<\tau)(\theta_i < \theta_j)(\delta_j)},\end{equation} 
where $\hat{G}(\cdot)$ is the Kaplan-Meier estimate for the censoring distribution. Due to these coefficients, $U_C$ converges to a quantity that is independent of the censoring distribution. $U_C$ takes values between 0 and 1, with higher values indicating a better fit.


It is important to note that the metrics described above are not specifically designed for survival trees, and therefore have certain limitations when applied in this context. The Cox partial likelihood score and the C-statistics  become less informative when a large number of observations have the same predicted risk coefficient, which is generally the case in tree models. Increasing the number of nodes in the tree may inflate these scores even if the overall quality of the model does not improve. 

\bigskip

 \item \textbf{Integrated Brier score}

The Brier score metric is commonly used to evaluate classification trees \citep{brier1950verification}. It was originally developed to verify the accuracy of a probability forecast, primarily purposed for weather forecasting. The most common formula calculates the mean squared prediction error:

\begin{equation} B = \frac{1}{n}\sum_{i}^n(\hat{p}(y_i) - y_i)^2,\end{equation} 

where $n$ is the sample size,  $y_i \in \{0,1\}$ is the outcome of observation $i$, and $\hat{p}(y_i)$ is the forecast probability of this observed outcome. In the context of survival analysis, the Brier score may be used to evaluate the accuracy of survival predictions at a particular point in time relative to the observed deaths at that time. We will refer to this as the Brier Point Score:

\begin{align} &BP_{\tau} = \frac{1}{|\mathcal{I}_{\tau}|}\sum_{i \in \mathcal{I}_{\tau}}(\hat{S_i}(\tau) - \mathbbm{1}(t_i >\tau))^2, \\ \text{ where } & \mathcal{I}_{\tau} = \{i\in \{1,\dots, n\},| t_i \geq \tau \text{ or } \delta_i=1\}. \end{align} 
In this case, $\hat{S_i}(\tau)$ is the predicted survival probability for observation $i$ at time $\tau$ and $\mathcal{I}_{\tau}$ is the set of observations that are known to be alive/dead at time $\tau$. Observations censored before time $\tau$ are excluded from this score, as their survival status is unknown. 

Applying this version of the Brier score may be useful in applications where the main outcome of interest is survival at a particular time, such as the 1-year survival rates after the onset of a disease.  In the experiments that follow, the point-wise Brier Score will be evaluated at the median observation time in each dataset. For easy interpretation, the reported scores are normalized relative to the score for a null model, i.e.
\begin{equation} BPR_{\tau}=1-\frac{ BP_{\tau}(T)}{ BP_{\tau}({T^0})}.\end{equation} 

The Brier Point score has two significant disadvantages in survival analysis. First, it assess predictive accuracy of survival models a single point in time rather than over the entire observation period, which is not well-suited to applications where survival distributions are the outcome of interest. Second, it becomes less informative as the number of censored observations increases, because a greater number of observations are discarded when calculating the score.

\citet{graf1999assessment} have addressed these challenges by proposing an adjusted version of the Brier Score for survival datasets with censored outcomes. Rather than measuring the accuracy of survival predictions at a single point, this measure aggregates the Brier score over the entire time interval observed in the data.  This modified measure is commonly used in the survival literature and has been interchangeably called the Brier Score or the Integrated Brier Score by various authors \citep{btn265}. In this paper, we will refer to the metric specific to survival analysis as the Integrated Brier score (IB), defined as 
\begin{align} IB = \frac{1}{n}\frac{1}{t_{max}}\sum_{i=1}^n\int_0^{t_i} \frac{(1-\hat{S}_{i}(t))^2}{\hat{G}_{i}(t)} dt  +
\delta_i\int_{t_i}^{t_{max}} \frac{(\hat{S}_{i}(t))^2}{\hat{G}_{i}(t_i)} dt.\end{align}

The IB score uses Kaplan-Meier estimates for both the survival distribution, $\hat{S}(t)$, and the censoring distribution, $\hat{G}(t)$. In a survival tree model, these estimates are obtained by pooling observations in each node in the tree, i.e., $\hat{S}_i(t)=\hat{S}_{T(i)}(t)$.
The IB score is a weighted version of the original Brier Score, with the weights being $1/\hat{G}_{i}(t_i)$ if an event occurs before time $t_i$, and $1/\hat{G}_{i}(t)$ if the event occurs after time t. 

We report the Integrated Brier score ratio (IBR), which compares the sum of the Integrated Brier scores in a given tree to the corresponding Integrated Brier scores in a null tree\footnote{\citet{radespiel2003comparison} calls this \textit{explained residual variation }}:
\begin{equation} IBR=1-\frac{ IB(T)}{ IB({T^0})}.\end{equation} 
\end{enumerate}

We note that all of the above metrics have some limitations and do not provide definitive evidence that one model is better than another. In practice, these metrics often provide contradictory assessments when comparing different tree models. For example, our empirical experiments comparing three candidate models were only able to identify a non-dominated model for about 30\% of the instances. In the other 70\% of our test cases, none of the three candidate models scored at least as high as the other models on all metrics.
 
These limitations make it difficult to obtain an unambiguous comparison between the performance of different survival tree algorithms. To address this challenge, we will now introduce a simulation procedure and associated accuracy metrics that are specifically designed to assess survival tree models.

\subsection{Simulation accuracy metrics}
\label{sec:simulationmetrics}
%
A key difficulty in selecting performance metrics for survival tree models is that the definition of ``accuracy'' can depend on the context in which the model will be used. 
For example, consider a survival tree  that models the relationship between lifestyle factors and age of death. A medical researcher may use such a model to \emph{identify risk factors} associated with early death, while an insurance firm may use this model to \emph{predict mortality risks} for individual clients in order to estimate the volume of life insurance policy pay-outs in the coming years. The medical researcher is primarily interested whether the model has identified important splits, while the insurer is more focused on whether the model can accurately estimate survival distributions.

In subsequent sections we refer to these two properties as \textit{classification accuracy} and \textit{prediction accuracy}. We develop metrics to measure these outcomes in simulated datasets with the following structure:

Let $i=1,\dots,n$ be a set of observations  with independent, identically distributed covariates $\mathbf{X}_{i}=(X_{ij})_{j=1}^m$. Let $C$ be a tree model that partitions observations based on these covariates such that $C_i = C(\mathbf{X}_{i})$ is the index of the leaf node in $C$ that contains individual $i$. Let $S_i$ be a random variable representing the survival time of observation $i$, with distribution $S_i\sim F_{C_i}(t)$. The survival distribution of each individual is entirely determined by its location in the tree $C$, and so we refer to $C$ as the ``true'' tree model. 

 This underlying tree structure provides an unambiguous target against which we can measure the performance of empirical survival tree models. In this context, an empirical survival tree model $T$ has high accuracy if it achieves the following objectives:\begin{enumerate}
\item Classification accuracy: the model recovers structure of the true tree (i.e., $T(\mathbf{X}_{i})=C(\mathbf{X}_{i})$). 
\item Prediction accuracy: the model recovers the corresponding survival distributions of the true tree (i.e., $\hat{F}_{T_i}(t)={F}_{C_i}(t)$). \end{enumerate}
It is important to recognize that these two objectives are not necessarily consistent, particularly in small samples. Trees with perfect classification accuracy may have a small number of observations in each leaf node, leading to noisy survival estimates with low prediction accuracy. 

\subsection{Classification accuracy metrics}\label{sec_classification}
\begin{figure}
\centering
\includegraphics[width=0.9\textwidth]{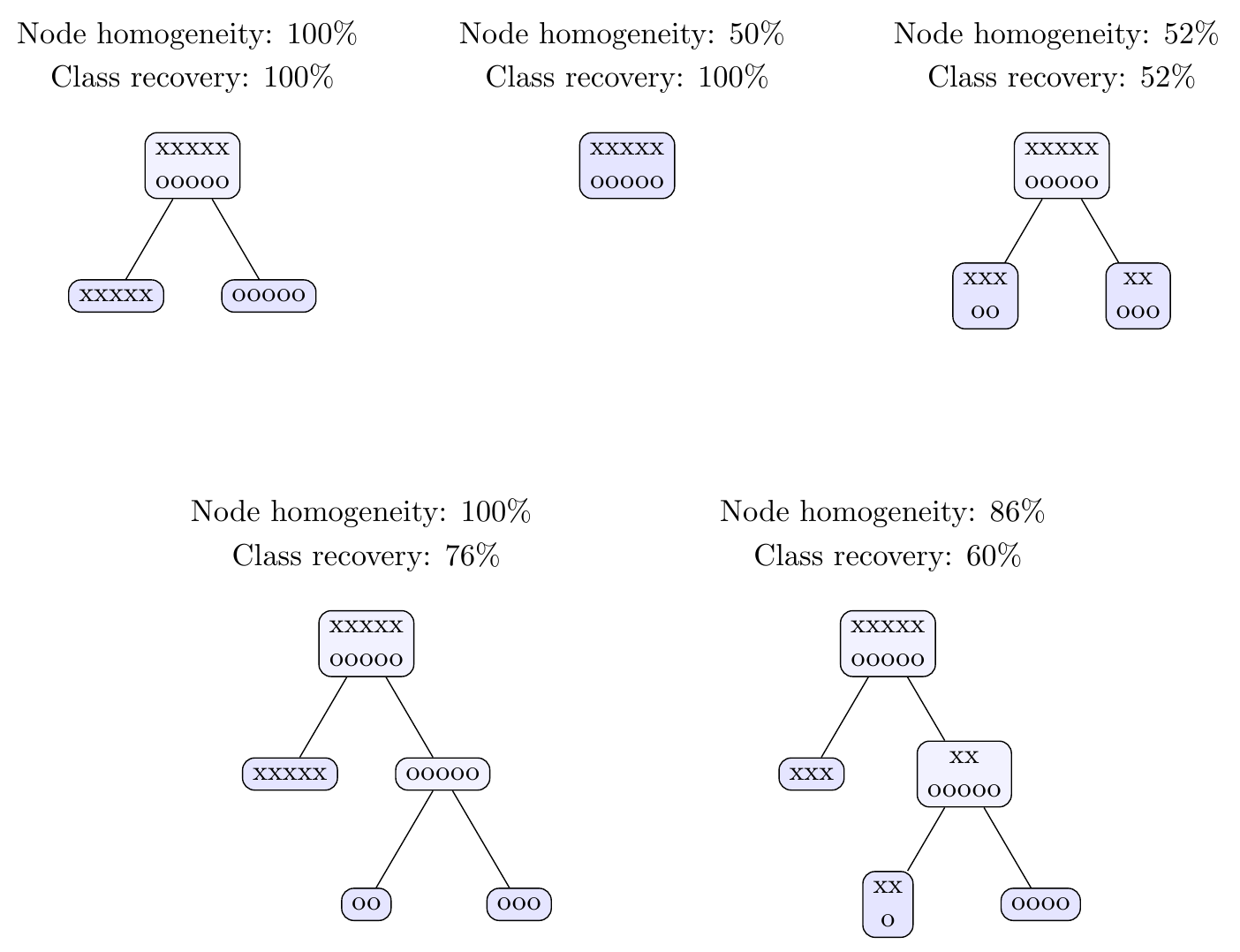}
\caption{Classification accuracy metrics for a survival tree with two classes of observations. The top left tree represents the true tree model.}\label{fig:classification}
\end{figure}

We measure the classification accuracy of an empirical tree model ($T$) relative to the true tree ($C$) using the following metrics:
\begin{enumerate}
\item \textbf{Node homogeneity}
The node homogeneity statistic measures the proportion of the observations in each node $k\in T$ that have the same true class in $C$. Let $p_{k,l}$ be the proportion of observations in node $k \in T$ that came from class $\ell \in C$ and let $n_{k,l}$ be the total number of observations at node $k \in T$ from class $\ell \in C$. Then,
\begin{equation} NH = 
\frac{1}{n}\sum_{k \in T}\sum_{l \in C} n_{k,l}p_{k,l}.\end{equation}
A score of $NH = 1$ indicates that each node in the new tree model contains observations from a single class in $C$. This does not necessarily mean that the structure of $T$ is identical to $C$ --- For example, a saturated tree with a single observation in each node would have a perfect node homogeneity score (see Figure~\ref{fig:classification}). The node homogeneity metric is therefore biased towards larger tree models with few observations in each node.

\item \textbf{Class recovery}

Class recovery is a measure of how well a new tree model is able to keep similar observations together in the same node, thereby avoiding unnecessary splits. Class recovery is calculated by counting the proportion of observations from a true class $\ell \in C$ that are placed in the same node in $T$. Let $q_{k,l}$ be the proportion of observations from class $\ell \in C$ that are classified in node $k \in T$  and let $n_{k,l}$ be the total number of observations at node $k \in T$ from class $\ell \in C$. Then,
\begin{equation}  CR = 
\frac{1}{n}\sum_{\ell \in C}\sum_{k \in T} n_{k,l}q_{k,l}.\end{equation} 

This metric is biased towards smaller trees, since a null tree with a single node would have a perfect class recovery score. It is therefore useful to consider both the class recovery and node homogeneity scores simultaneously in order to assess the performance of a tree model (see Figure~\ref{fig:classification} for examples). When used together, these metrics indicate how well the model $T$ reflects the structure of the true model $C$.
\end{enumerate}

The node homogeneity and class recovery scores can also be used to compare any two tree models, $T_1$ and $T_2$. In this case, these metrics should be interpreted as a measure of structural similarity between the two tree models.  Note that when $T_1$ and $T_2$ are applied to the same dataset,
 the node homogeneity for model $T_1$ relative to $T_2$ is equivalent to the class recovery for $T_2$ relative to $T_1$, and vice versa. The average node homogeneity score for $T_1$ and $T_2$ is therefore equal to the average class recovery score for $T_1$ and $T_2$. We will refer to this as the \textit{similarity score} for models $T_1$ and $T_2$. 

\subsection{Prediction accuracy metric}
Our prediction accuracy metric measures how well the non-parametric Kaplan-Meier curves at each leaf in $T$ estimate  true the survival distribution of each observation.

\begin{enumerate}
\item \textbf{Area between curves (ABC)}

For an observation $i$ with true survival distribution $F_{C_i}(t)$, suppose that $\hat{S}_{T_i}(t)$ is the Kaplan-Meier estimate at the corresponding node in tree $T$ (see Figure~\ref{fig:ABC}). The area between the true survival curve and the tree estimate is given by
\begin{equation} ABC_i^T = \frac{1}{t_{max}}\int_{0}^{t_{max}} |1-F_{C_i}(t)-\hat{S}_{T_i}(t)|dt.\end{equation}  To make this metric easier to interpret, we compare the area between curves in a given tree to the score of a null tree with a single node ($T_0$). The area ratio (AR) is given by \begin{equation}AR=1-\frac{\sum_i ABC_i^T}{\sum_i ABC_i^{T^0}}.\end{equation} Similar to the popular $R^2$ metric for regression models, the AR indicates how much accuracy is gained by using the Kaplan-Meier estimates generated by the tree relative to the baseline accuracy obtained by using a single estimate for the whole population.  

\begin{figure}
\centering
\includegraphics[width=0.6\textwidth]{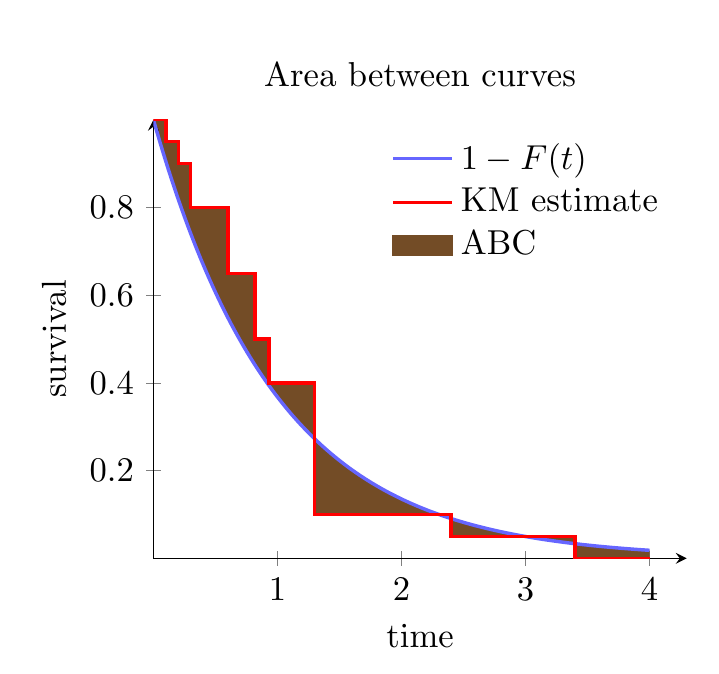}
\caption{An illustration of the area between the true survival distribution and the Kaplan-Meier curve.}
\label{fig:ABC}
\end{figure}

\end{enumerate}

\section{Simulation results}\label{sec:simulations}
In this section we evaluate the performance of the Optimal Survival Trees (OST) algorithm and compare it to two existing survival tree models available in the \pkg{R} packages \pkg{rpart} and \pkg{ctree}. Our tests are performed on simulated datasets with the structure described in Section~\ref{sec:simulationmetrics}.

\subsection{Simulation procedure}\label{sec:simulationparam}
The procedure for generating simulated datasets in these experiments is as follows:

\begin{enumerate}
\item Randomly generate a sample of 20000 observations with six covariates. The first three covariates are uniformly distributed on the interval $[0,1]$ and remaining three covariates are discrete uniform random variables with 2, 3 and 5 levels.
\item  Generate a random ``ground truth'' tree model, $C$, that partitions the dataset based on these six covariates  (see Algorithm~\ref{alg:growtree} in the Appendix).
\item Assign a survival distribution to each leaf node in the tree $C$ (see Appendix for a list of distributions). 
\item Classify observations into node classes $C_i = C(\mathbf{X}_i)$ according to the ground truth model. Generate a survival time, $s_i$, for each observation based the survival distribution of its node: $S_i\sim F_{C_i}(t)$.
\item Generate a censoring time for each observation, $c_i = \kappa(1-u_i^2)$, where $u_i$ follows a uniform distribution and $\kappa$ is a non-negative parameter used to control the proportion of censored individuals.
\item Assign observation times $t_i=\min(s_i,c_i)$. Individuals are marked as censored ($\delta_i=0$) if $t_i=c_i$. 
\end{enumerate}

We used this procedure to generate 1000 datasets based on ground truth trees with a minimum depth of 3 and a maximum depth of 4 (i.e., $2^4=16$ leaf nodes). In each instance, 10000 observations were set aside for testing the tree models. Training datasets of $n$ observations were sampled from the remaining data for $n \in \{100,200,500,1000,2000,5000,10000\}$.

In addition to varying the size of the training dataset, we also varied the proportion of censored observations in the data by adjusting the parameter $\kappa$. Censoring was applied  at nine different levels to generate examples with low censoring~(0\%,~10\%,~20\%), moderate censoring~(30\%,~40\%,~50\%) and high censoring~(60\%,~70\%,~80\%). In total, 63 OST models were trained for each dataset to test each of the seven training sample sizes at each of the  nine censoring levels.

We evaluated the performance of the OST algorithm relative to two existing survival tree algorithms available in the  \pkg{R} packages \pkg{rpart} \citep{therneau2010rpart} and \pkg{ctree} \citep{hothorn2010party}. Each of the three algorithms was trained and tested on exactly the same data in each instance.

Each of the three algorithms tested require two input parameters that control the model size: a maximum tree depth and a complexity/significance parameter that determines which splits are worth keeping in the tree (the interpretation of the \pkg{ctree} significance parameter is different to the complexity parameters in the OST and \pkg{rpart} algorithms, but it serves a similar function).

Since neither \pkg{rpart} nor \pkg{ctree} have built-in methods for selecting tree parameters, we used a similar 5-fold cross-validation procedure to select the parameters for each algorithm.  We considered tree depths up to three levels greater than the true tree depth and complexity parameter/significance values between 0.001 and 0.1  for the \pkg{rpart} and \pkg{ctree} algorithms (the OST complexity parameter is automatically selected during training). Equation~(\ref{eqn:error}) was used as the scoring metric to evaluate out-of-sample performance during cross-validation, and the minimum node size for all algorithms was fixed at 5 observations. 

\subsection{Results}
\begin{figure}[b]
\includegraphics[width = \linewidth]{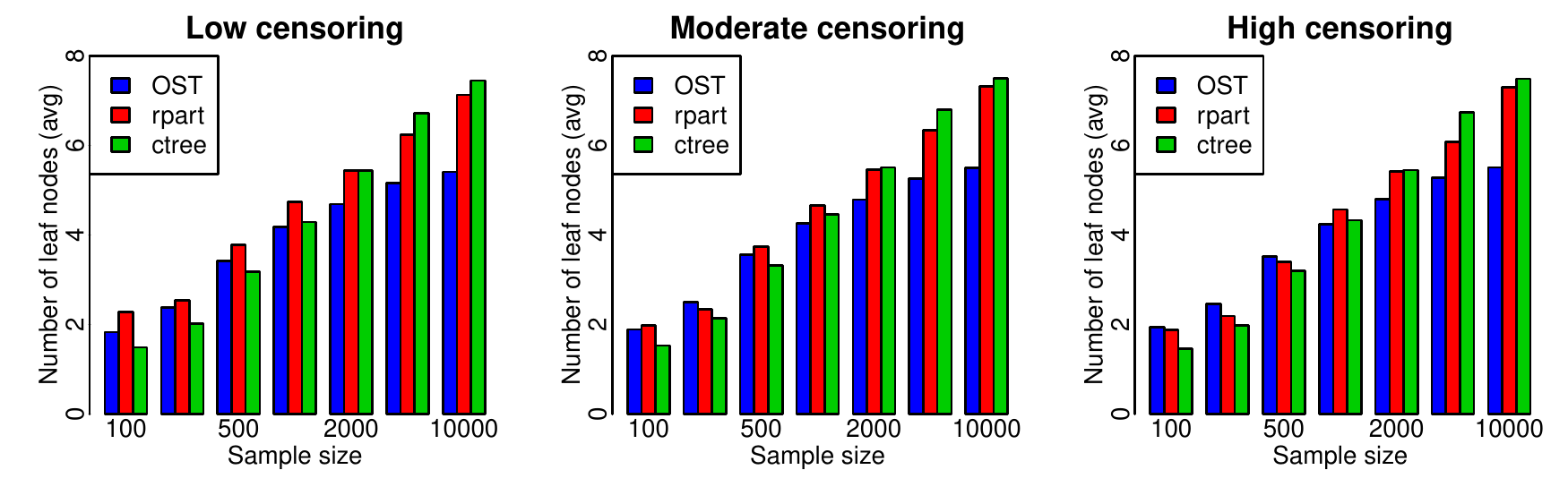}
\caption{The average tree size for models trained on various sample sizes.}\label{fig:nodes}
\end{figure}
To demonstrate the effect of this cross-validation procedure, we summarize the average size of the models produced by each algorithm in Figure~\ref{fig:nodes}.  We see a clear link between tree size and the number of training observations, indicating the cross-validation procedure is selecting more conservative depth/complexity parameters when relatively little data is available. In larger datasets, the OST  models grow to approximately the same size as the true tree models (6 nodes, on average), while the 
\pkg{rpart} and \pkg{ctree} models models are slightly larger.

\subsubsection{Survival analysis metrics}
\begin{figure}
\includegraphics[width = \linewidth]{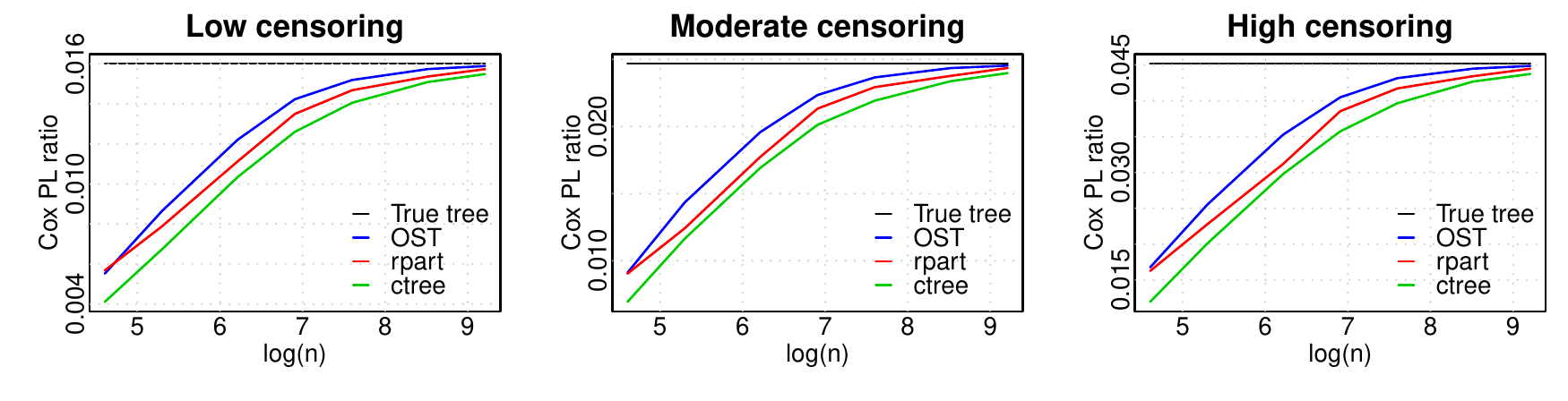}
\includegraphics[width = \linewidth]{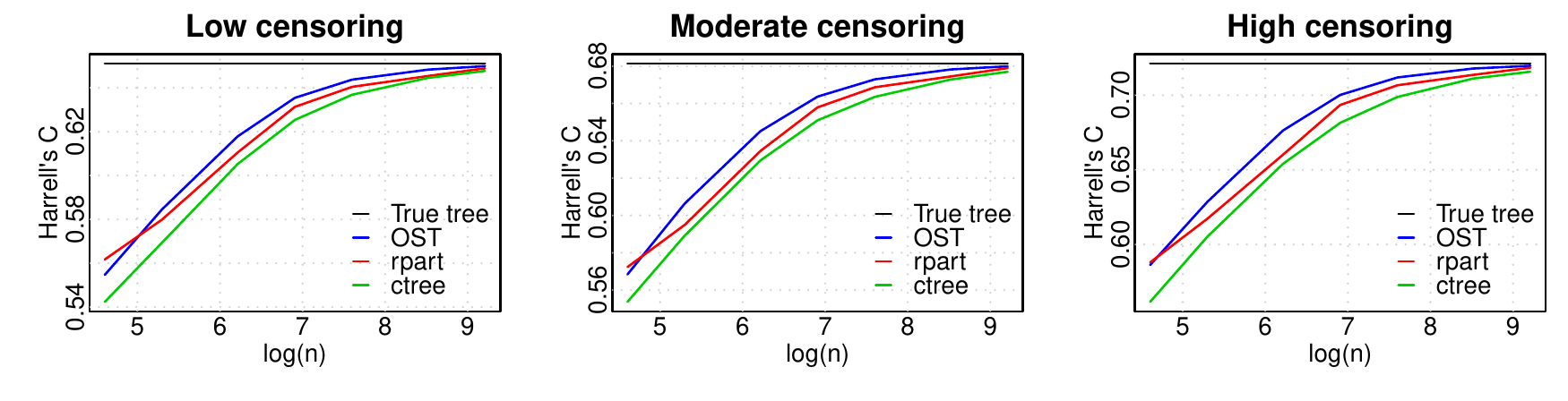}
\includegraphics[width = \linewidth]{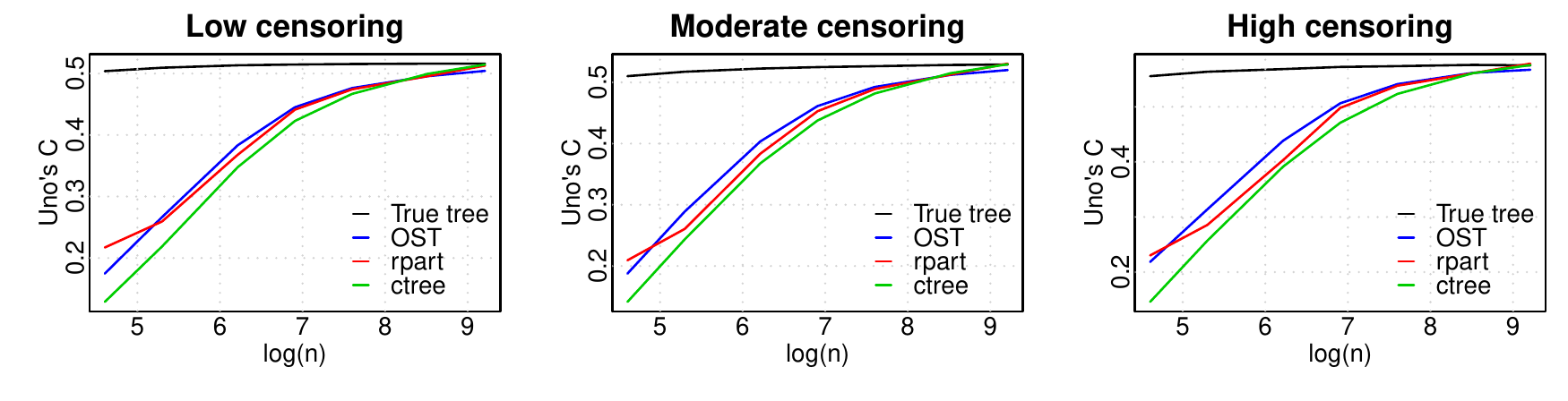}
\includegraphics[width = \linewidth]{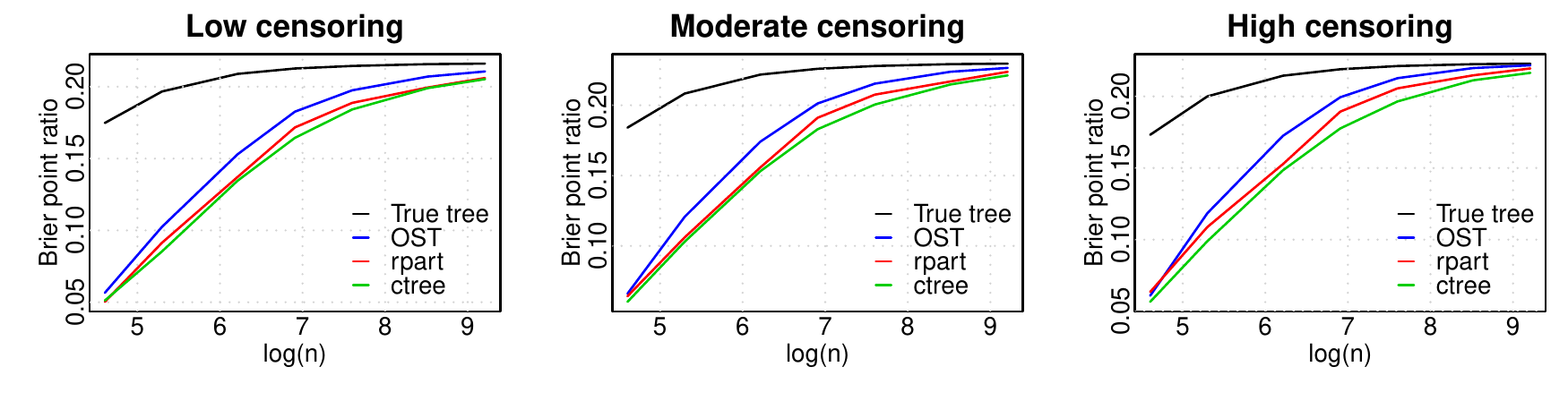}
\includegraphics[width = \linewidth]{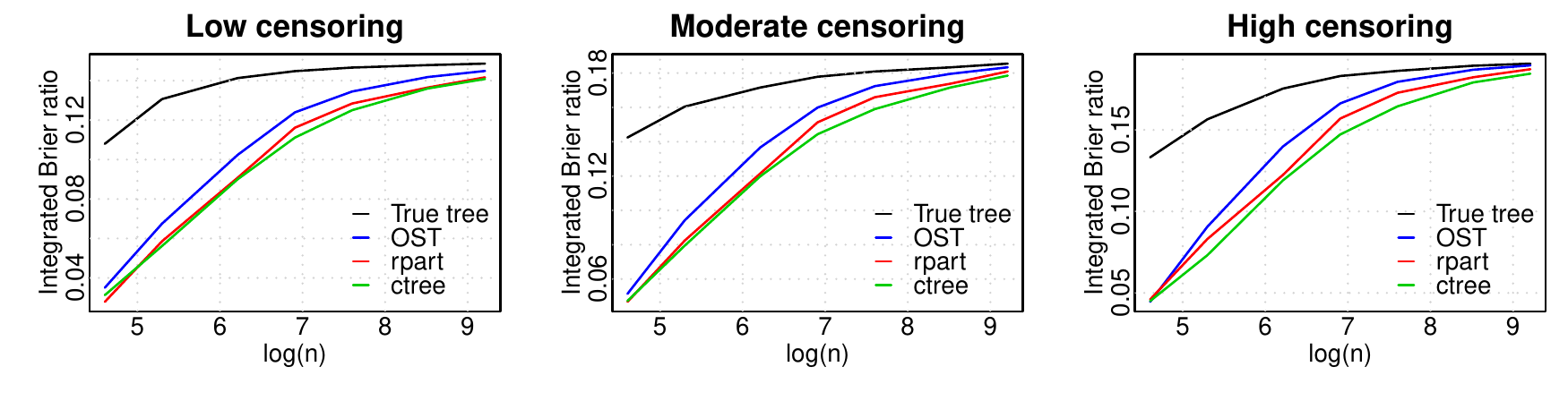}
\caption{A summary of the survival model metrics from simulation experiments. The average out-of-sample outcomes for each algorithm are shown in color, while the performance of the true tree model, $C$, in indicated in black. }\label{fig:survival_metrics}
\end{figure}

Figure~\ref{fig:survival_metrics} summarizes the performance of each algorithm in our simulations using the  four survival model metrics from Section~\ref{sec:metrics}. The values displayed in each chart are the average out-of-sample performance statistics across all test instances. 

As expected, the average performance of all three algorithms consistently improves as the size of the training dataset increases. The performance statistics also increase as the proportion of censored observations increases, which seems counter-intuitive (we would expect more censoring to lead to less accurate models). In the case of the Cox partial likelihood and C-statistics, this trend is directly linked to the number of observed deaths, since only observations with observed deaths contribute to the partial likelihood and concordance scores. Similarly, censored observations do not contribute to the Integrated Brier Score after their censoring time.

Each chart also indicates the performance of the true tree model, $C$, as a point of comparison for the other algorithms. The true tree model performs significantly better than the empirical models trained on smaller datasets, but all three algorithms approach the performance of the true tree for very large sample sizes.

Based on these results, we conclude that the average performance of the OST algorithm in these simulations is consistently better than either of the other two algorithms. In order to understand why this algorithm is able to generate better models, we now analyse the results of the tree metrics introduced in Section~\ref{sec:simulationmetrics}.

\subsubsection{Classification accuracy}

\begin{figure}
\includegraphics[width = \linewidth]{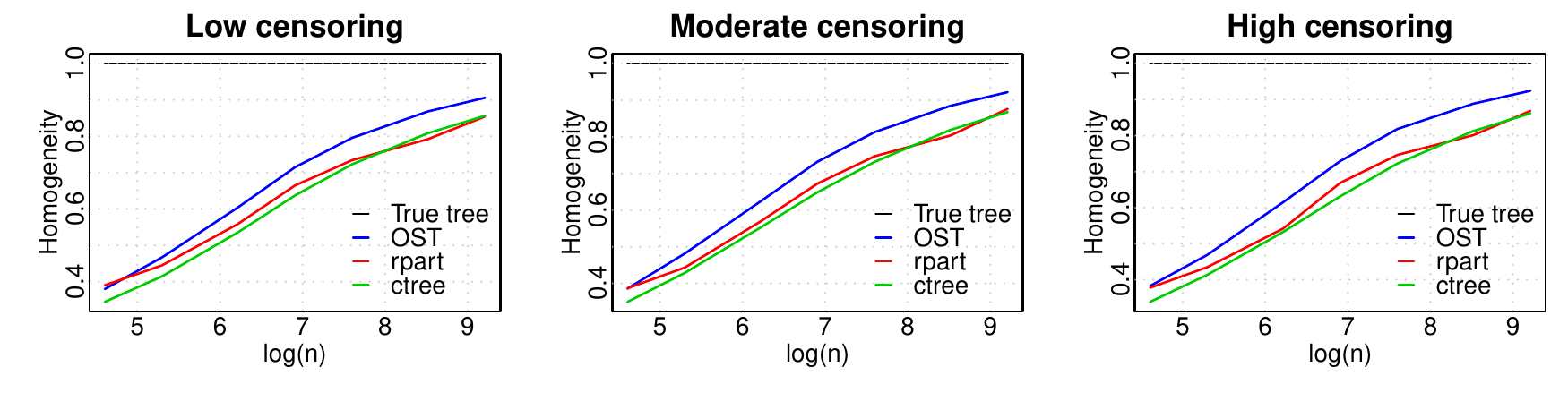}
\includegraphics[width = \linewidth]{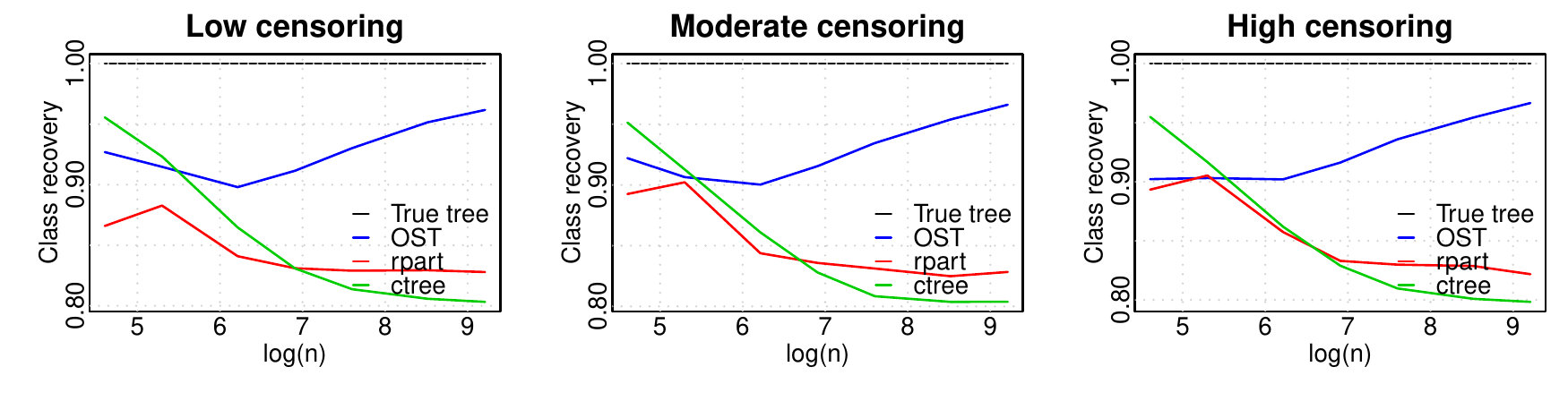}
\caption{A summary of the classification accuracy metrics for survival tree algorithms.}\label{fig:homog_sep}
\end{figure}

The out-of-sample classification accuracy metrics for all three algorithms are summarized in Table~\ref{tab:Detresults} and Figure~\ref{fig:homog_sep}. The average node homogeneity/class recovery scores are given side-by-side to allow for a comprehensive assessment of each algorithm's performance. These results confirm that the OST models perform significantly better than the other two models across all censoring levels.
\setlength{\tabcolsep}{4.5pt}
\begin{table}
\centering
\footnotesize
\begin{tabular}{crrrrrrrrr}
\toprule
 & \multicolumn{3}{c}{ Low censoring}&  \multicolumn{3}{c}{ Moderate censoring} & \multicolumn{3}{c}{ High censoring}\\
\midrule
$n$ &\hspace{3pt}  rpart & ctree & OST  & \hspace{8pt} rpart & ctree & OST  & \hspace{8pt} rpart & ctree & OST \\
\midrule
100 & 38/87 & \textbf{40}/77 & 37/\textbf{93} & 38/90 & \textbf{40}/78 & 37/\textbf{92} & 37/89 & \textbf{40}/78 & 37/\textbf{90}  \\ 200 & 42/89 & \textbf{45}/76 & 43/\textbf{91} & 42/\textbf{90} & \textbf{46}/77 & 45/90 & 42/\textbf{91} & 45/78 & \textbf{45}/90  \\ 500 & 53/84 & 56/71 & \textbf{57}/\textbf{88} & 55/84 & 57/70 & \textbf{59}/\textbf{88} & 53/85 & 56/72 & \textbf{59}/\textbf{88}  \\ 1000 & 63/82 & 66/63 & \textbf{68}/\textbf{89} & 65/82 & 67/63 & \textbf{70}/\textbf{89} & 64/82 & 66/64 & \textbf{70}/\textbf{89}  \\ 2000 & 70/81 & 73/57 & \textbf{76}/\textbf{89} & 72/81 & 75/57 & \textbf{78}/\textbf{90} & 72/81 & 74/58 & \textbf{78}/\textbf{90}  \\ 5000 & 76/80 & 82/53 & \textbf{84}/\textbf{91} & 77/80 & 83/53 & \textbf{85}/\textbf{92} & 77/80 & 82/53 & \textbf{85}/\textbf{91}  \\ 10000 & 82/79 & 85/50 & \textbf{87}/\textbf{91} & 84/79 & 86/51 & \textbf{89}/\textbf{92} & 84/78 & 86/51 & \textbf{88}/\textbf{91}  \\
\bottomrule

\end{tabular}
\caption{A summary of the average node homogeneity/class recovery scores for synthetic experiments.}
\label{tab:Detresults}
\end{table}

The node homogeneity scores for all three algorithms increase with larger sample sizes, indicating that the availability of additional data leads to better detection of relevant splits. In large populations, the OST algorithm selects more efficient splits than the other models and is able to achieve better node homogeneity with fewer splits (recall Figure~\ref{fig:nodes} --- the OST models trained on large data sets have fewer leaf nodes than the other models, on average).  

The relationship between tree size and class recovery rates is somewhat more complicated. In datasets smaller than 500 observations the class recovery rates seem to be closely linked to the tree size: the \pkg{ctree} models have the highest average class recovery for models trained on 100 and 200 observations, and also the smallest number of nodes (see Figure~\ref{fig:nodes}). However, this trend does not hold in datasets with 500 observations, where OST models are larger than the \pkg{ctree} models on average, but also have slightly better class recovery. This suggests that tree size is no longer a dominant factor in larger datasets ($n\geq 500$).

In these larger datasets we observe distinct trends in class recovery scores. The OST class recovery rate increases consistently despite the increases in model size, which means that the OST models are able to produce more complex trees without overfitting in the training data. By contrast, both of the other algorithms have consistently worse class recovery rates as sample size increases and their models become larger. Based on this trend, neither of these algorithms will reliably converge to the true tree.

\subsubsection{Prediction accuracy}

\begin{table}
\centering
\footnotesize
\begin{tabular}{crrrrrrrrr}
\toprule
 & \multicolumn{3}{c}{ Low censoring}&  \multicolumn{3}{c}{ Moderate censoring} & \multicolumn{3}{c}{ High censoring}\\
\midrule
$n$ &\hspace{3pt}  rpart & ctree & OST  & \hspace{15pt}  rpart & ctree & OST  & \hspace{15pt}  rpart & ctree & OST \\
\midrule
 100 & 6.87 & 4.79 & \textbf{9.30} & 10.61 & 7.74 & \textbf{11.01} & \textbf{10.79} & 7.76 & 9.99  \\ 200 & 18.69 & 16.82 & \textbf{20.99} & 21.93 & 21.09 & \textbf{25.25} & 24.20 & 21.24 & \textbf{26.13}  \\ 500 & 35.03 & 32.56 & \textbf{41.17} & 40.14 & 37.12 & \textbf{47.16} & 40.84 & 38.34 & \textbf{48.21}  \\ 1000 & 51.27 & 44.29 & \textbf{56.44} & 57.28 & 49.68 & \textbf{61.99} & 58.86 & 51.30 & \textbf{63.95}  \\ 2000 & 62.76 & 55.04 & \textbf{67.97} & 68.71 & 60.30 & \textbf{73.53} & 70.35 & 61.67 & \textbf{75.31}  \\ 5000 & 72.62 & 66.94 & \textbf{79.45} & 77.26 & 71.63 & \textbf{83.50} & 79.22 & 72.38 & \textbf{84.68}  \\ 10000 & 80.06 & 73.57 & \textbf{84.41} & 84.84 & 77.44 & \textbf{87.77} & 85.80 & 77.94 & \textbf{88.72}  \\
 \bottomrule
\end{tabular}
\caption{A summary of the average Kaplan-Meier area ratio (AR) scores for simulation experiments.}
\label{tab:KMresults}
\end{table}

\begin{figure}
\includegraphics[width = \linewidth]{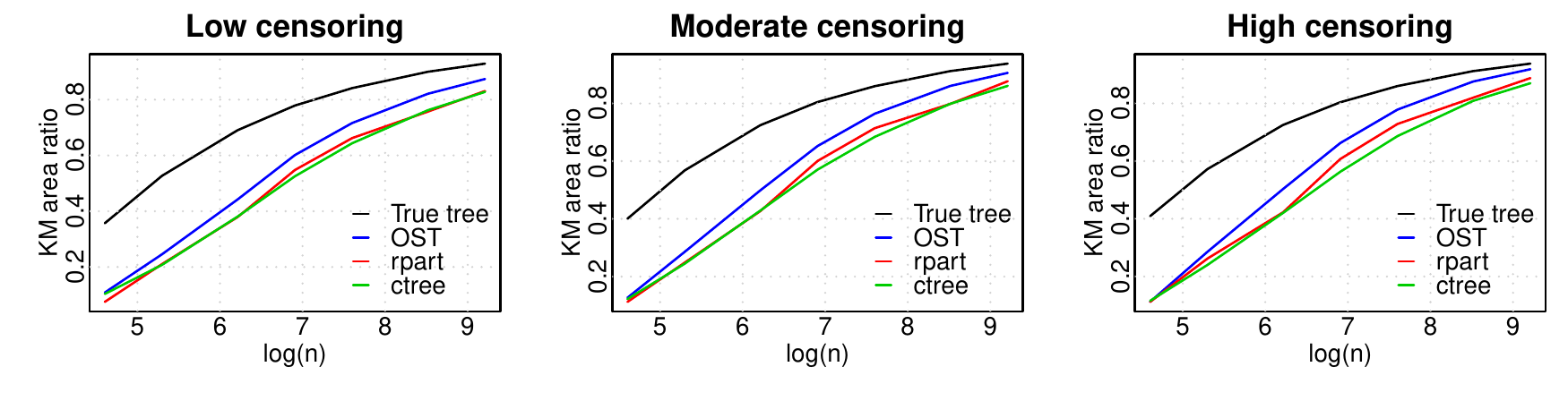}
\caption{A summary of the average Kaplan-Meier Area Ratio results for simulation experiments. The performance of the true tree model is indicated in black. }\label{fig:KM}
\end{figure}

The out-of-sample prediction accuracy metric for each of the three algorithms is summarized in Table~\ref{tab:KMresults} and Figure~\ref{fig:KM}. Overall, the results indicate that  sample size plays the most significant role in out-of-sample accuracy across all three algorithms. There is also a small increase in accuracy when censoring is increased, which is due to the reduction in the maximum observed time, $t_{max}$. 
The OST results are generally better than the other algorithms across all sample sizes, although the performance gap is relatively small in smaller datasets. 

To illustrate  the effect of sample size on the accuracy of the Kaplan-Meier estimates, Figure~\ref{fig:KM} also shows the curve accuracy metrics for the true tree, $C$.  It is immediately apparent that even the true tree models produce poor survival curve estimates in small datasets.
Based on these results, it may be necessary to increase the minimum node size to at least 50 observations in applications where Kaplan-Meier curves will be used to summarize survival tree nodes.

\subsubsection{Stability}
\begin{figure}
\includegraphics[width = \linewidth]{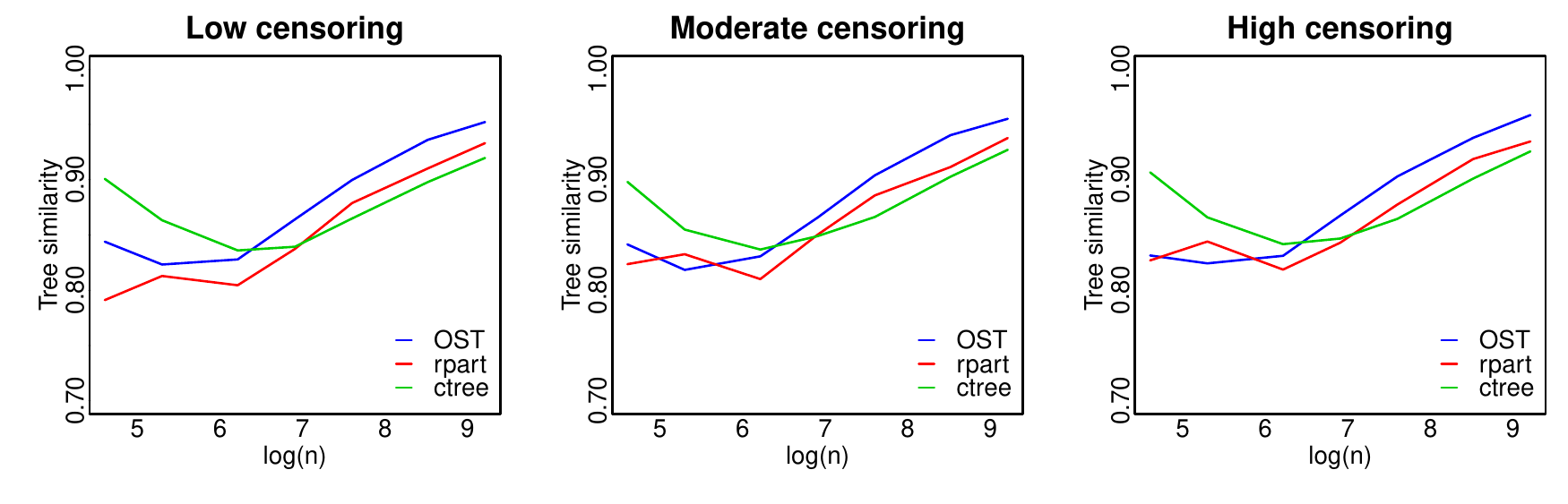}
\caption{A summary of the average similarity scores between pairs of trees trained on mutually exclusive sets of observations.}\label{fig:permute}
\end{figure}

\begin{figure}
\includegraphics[width = \linewidth]{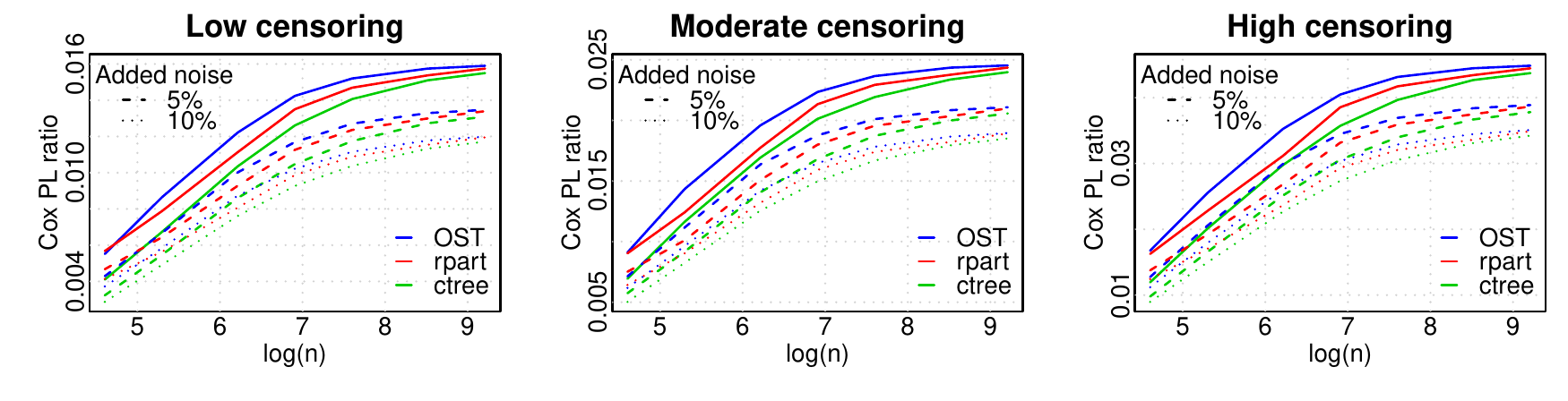}
\includegraphics[width = \linewidth]{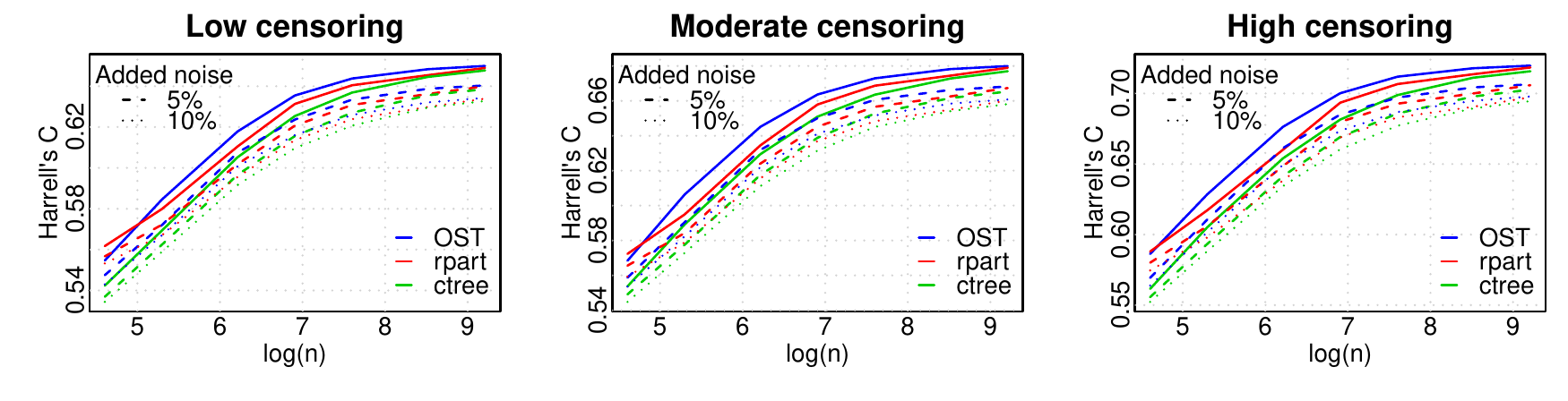}
\includegraphics[width = \linewidth]{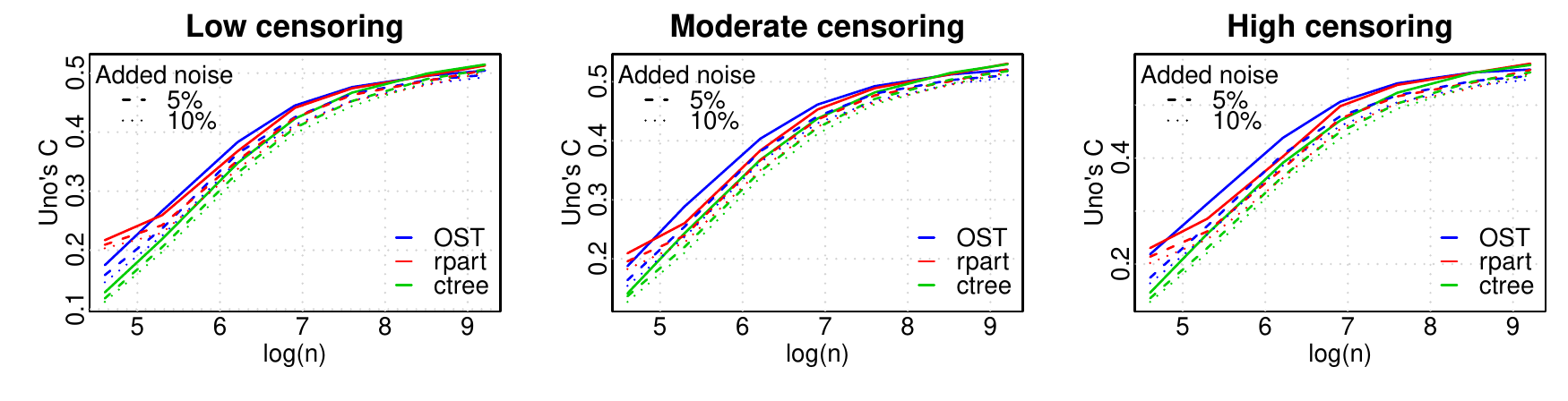}
\includegraphics[width = \linewidth]{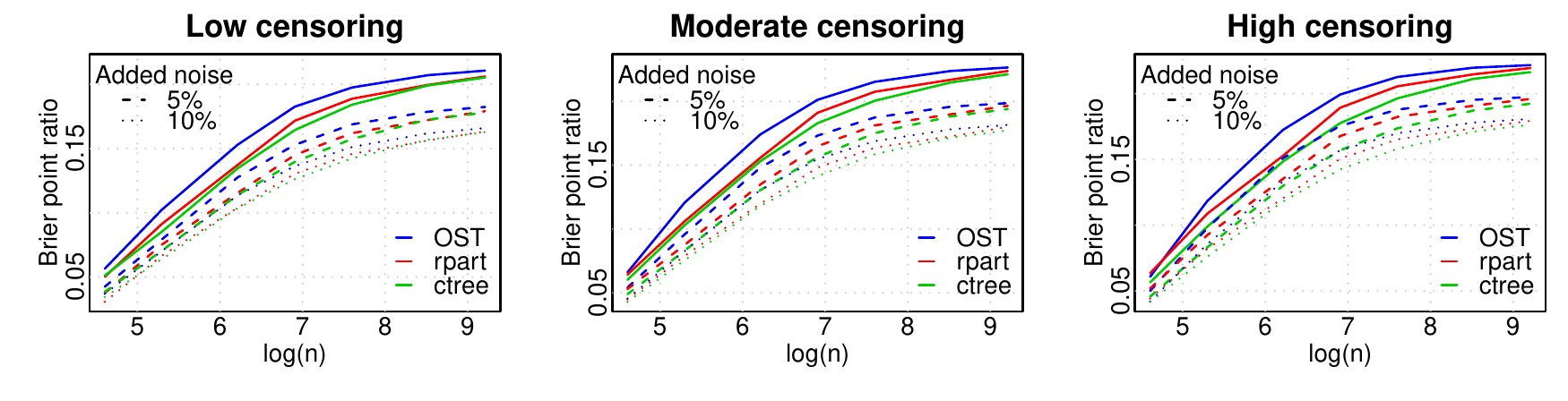}
\includegraphics[width = \linewidth]{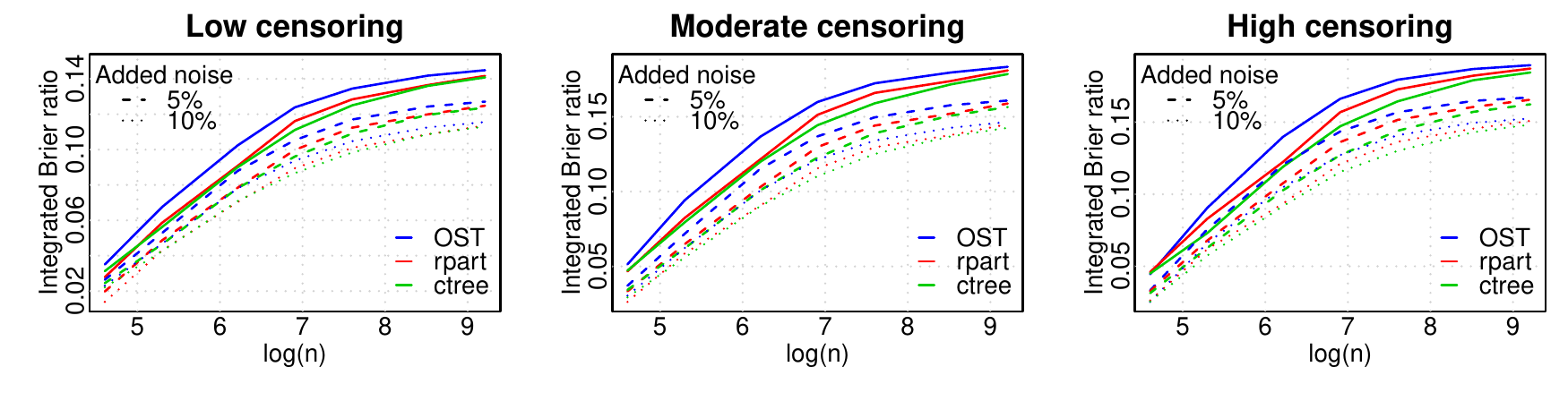}
\caption{A summary of survival tree accuracy metrics for datasets with added noise.}\label{fig:noise}
\end{figure}
\begin{figure}

\includegraphics[width = \linewidth]{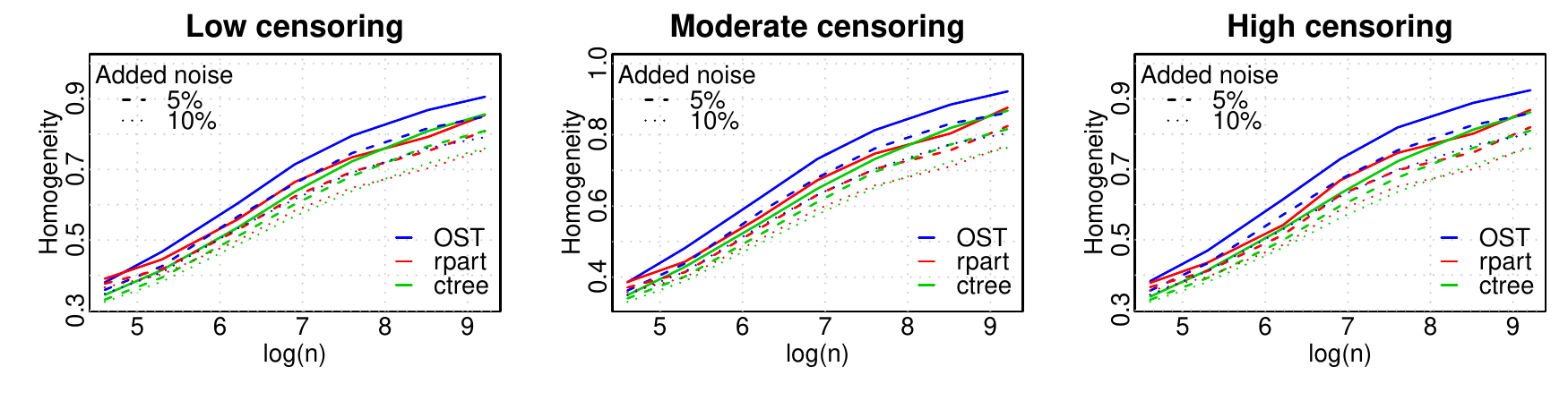}
\includegraphics[width = \linewidth]{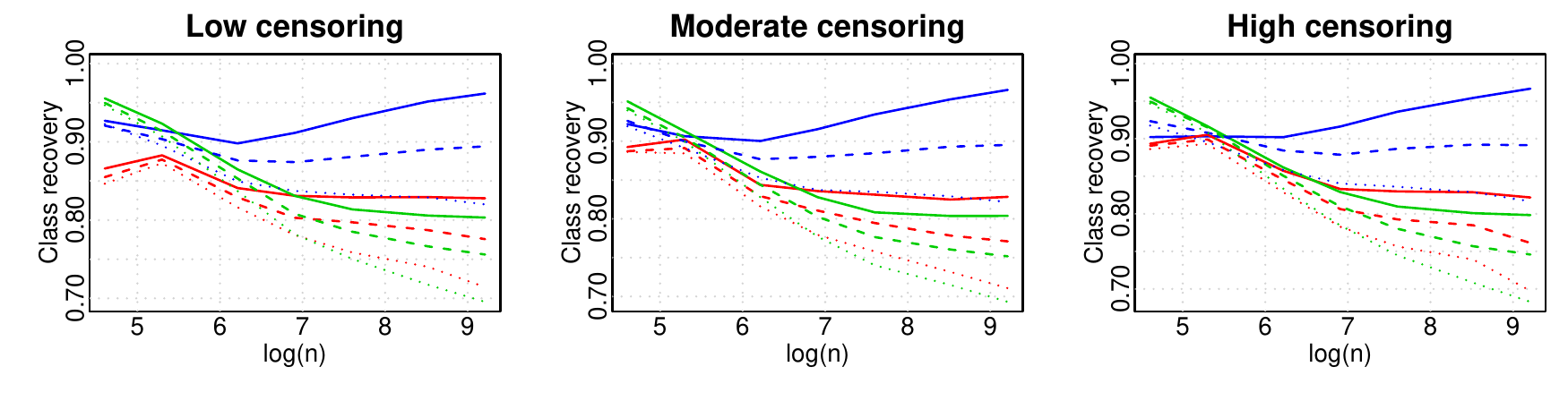}
\includegraphics[width = \linewidth]{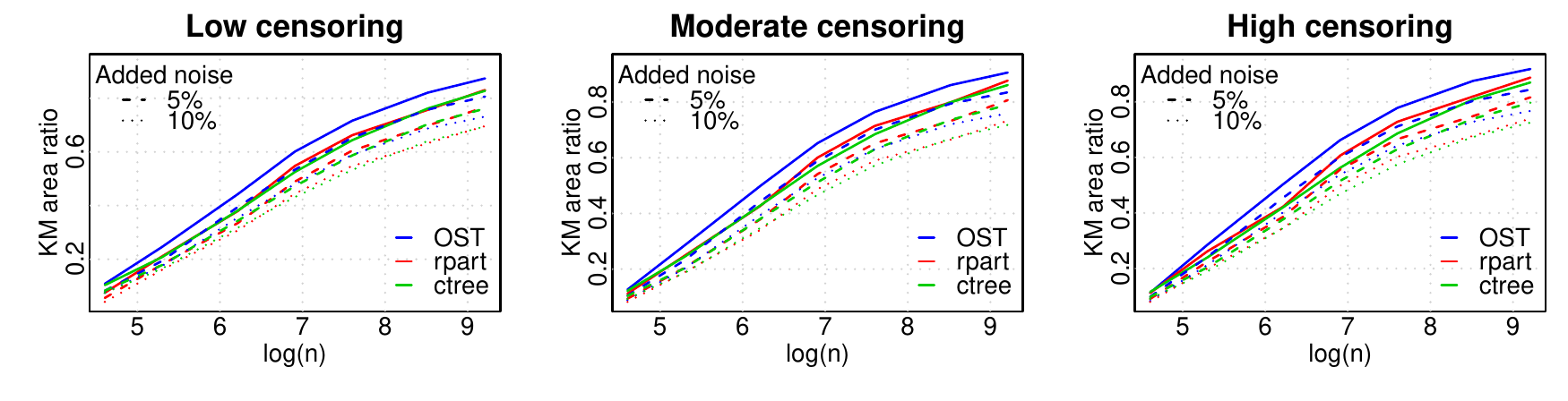}
\caption{A summary of simulation accuracy metrics for datasets with added noise.}\label{fig:noise2}
\end{figure}

 A frequent criticism of single-tree models is their sensitivity to small changes in the training data. This may be apparent  when a tree algorithm produces very different models for different training datasets sampled from the same population. This type of instability is often an indication that the model will not perform well on unseen data.
 
Given the challenges associated with measuring out-of-sample accuracy for survival tree algorithms, it may be tempting to use stability as a performance metric for these models. Stability is a necessary condition for accuracy in tree models (provided that a tree structure is suitable for the data) but stable models are not necessarily accurate. For example, greedy tree models with depth 1 may select the same split for all permutations of the training data, but these models will not be accurate if the data requires a tree of depth 3.

Although stability is not necessarily a good indicator of the quality of a model, it is nevertheless interesting to consider how the stability of globally optimized trees may differ to the stability of greedy trees. Globally optimized trees are theoretically capable of greater stability because they may include splits that are not necessarily locally optimal for a particular training dataset. However, globally optimized trees also consider a significantly larger number of possible tree configurations and  therefore have many more opportunities for overfitting on features of a particular training dataset.

 We ran two sets of experiments to investigate the stability of the  survival tree models in our simulations. In the first set of experiments we used each algorithm to train two models, $T_1$ and $T_2$, on completely separate training datasets of equal size. We then applied each model to the entire dataset (20000 observations) and used the tree similarity score described in Section~\ref{sec_classification} to assess the structural similarity between the two models. The average similarity scores for each algorithm are illustrated in Figure~\ref{fig:permute}.
 
 These results demonstrate that stability across different training datasets is not a sufficient condition for accuracy: models trained on 100 and 200 observations are both more stable and less accurate than models trained on 500 observations. The \pkg{ctree} algorithm produced the most stable results in smaller datasets due to the smaller model sizes selected during cross-validation. For example, 33.1\% of \pkg{ctree} models trained on 100 observations had fewer than 2 splits, compared to 29.5\% of the \pkg{rpart} models and 26.5\% of the OST models.

The stability results for larger training datasets ($n>1000$) are reasonably consistent with the accuracy metrics discussed above, and both stability and accuracy increase with sample size across all three algorithms. The OST models have the highest average similarity  scores in large datasets and the \pkg{rpart} models are slightly more stable than the \pkg{ctree} models.

 In the second set of stability experiments we investigated how small perturbations to the covariate values in the training dataset affect the out-of-sample accuracy of each model. We added noise to the training data by replacing the original continuous covariate values, $x_{ij}$, with ``noisy'' values $\tilde{x}_{ij}=x_{ij} + \epsilon_{ij}$. The initial covariates were uniformly distributed between 0 and 1 and the added noise terms were generated from the following two distributions:\begin{align*}
 &\epsilon_{ij}\sim U(-0.05,0.05) \quad \quad &&\text{(5\% noise), and }\\ &\epsilon_{ij}\sim U(-0.1,0.1)& &\text{(10\% noise).}
 \end{align*} A similar approach was applied to the categorical variables, which were generated by rounding off continuous values ($x_{ij}$ or $\tilde{x}_{ij}$) to the appropriate thresholds. Note that noise was only added to the observations used for training data; the testing data was unchanged.
 
 The results of these experiments are contrasted with the initial outcomes (without added noise) in Figures~\ref{fig:noise}-\ref{fig:noise2}. The effects of additional noise in the training data are visible in the results of all three algorithms and the drop in accuracy appears to be fairly consistent. Overall, the OST models maintain the highest scores regardless of noise.
 
These results indicate that perturbations in the training data affect the OST and greedy tree algorithms in similar ways. The OST algorithm's performance is diminished by adding noise to the training data, but its ability to consider a wider range of split configurations does not make it more sensitive to these perturbations. In fact, the OST algorithm is generally slightly more stable than the greedy algorithms across permutations of the training data because it tends to produce models that are consistently  closer to the true tree. 

\section{Computational results with real-world datasets}\label{sec:realdatasets}

In this section, we compare the performance of the OST, \pkg{rpart} and \pkg{ctree} algorithms on 44 real-world datasets. The datasets used for this analysis were sourced from the UCI repository \citep{UCI} and contained continuous outcome
measures.
 The selected datasets\footnote{We excluded the following types of datasets from our analysis: (1)  datasets used for time series predictions (multiple observations of each individual); (2) datasets with unclear variable definitions; (3) datasets which required significant  cleaning, pre-processing, or recoding; (4) datasets with too many variables ($p$) to cross-validate all three algorithms in reasonable times. Dataset selection was independent of the analysis of model accuracy.}  had sample sizes ranging from 63 observations to 100000, and the maximum number of features considered was 383.   

For each observation in these datasets, we generated censoring values $c_i = \kappa(1-u_i^2)$, where $u_i$ follows a uniform distribution. We adjusted the parameter $\kappa$ to generate different censoring levels (0\%,10\%,\dots,80\%) within each dataset. We then split each dataset into training and testing sets (50\%) and compared the performance of the three tree algorithms on each instance.

We applied the 5-fold cross-validation procedure described in Section~\ref{sec:simulationparam} to select the depth and complexity of each tree, allowing tree depths of up to 7 (128 leaf nodes). Both the OST and \pkg{ctree} algorithms produced trees with over 100 leaf nodes in some of the largest datasets, while the largest \pkg{rpart} trees had only 77 nodes. The smaller size of the \pkg{rpart} trees indicates that larger models performed poorly in the cross-validation step.

\begin{figure}
    \centering
    \includegraphics[width=\textwidth]{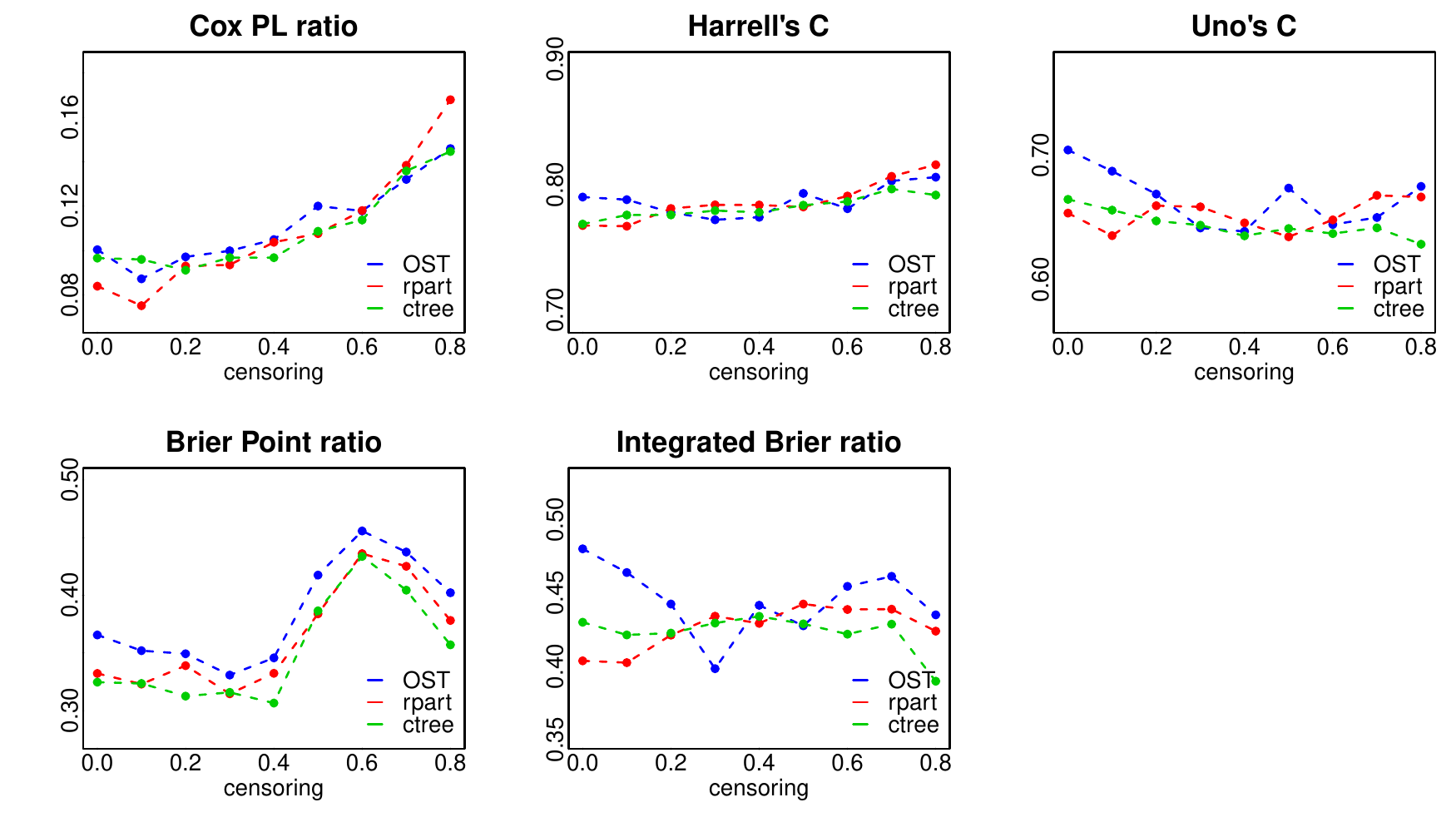}
    \caption{Average performance of survival tree models on real datasets with different levels of censoring.}
    \label{fig:Cens_real_world}
\end{figure}

On average, the OST models outperformed the other two algorithms across all 5 accuracy metrics. A summary of each algorithm's performance is given in Tables~\ref{tab:realresults}--\ref{tab:realresultsrank} and Figure~\ref{fig:Cens_real_world}, and aggregated results for each dataset are displayed in Table~\ref{tab:DataSpecific}. The difference in performance was not statistically significant for the Cox ratios and Harrell's C scores, where all three algorithms had very similar average outcomes, but OST models did score significantly better than the other algorithms on the remaining three metrics. OST models achieved the best score in 48-60\% of the instances tested, while the other algorithms each had undominated scores in 27-39\% of instances.

\begin{table}[h]
    \centering
    \begin{tabular}{l r rr r ll}
    \toprule
    & \multicolumn{3}{c}{Mean score} && \multicolumn{2}{c}{Paired T-Test $H_1$:} \\
    & OST & \pkg{rpart} & \pkg{ctree} & & $S_{OST}>S_{\pkg{rpart}}$ &  $S_{OST}>S_{\pkg{ctree}}$  \\ \cline{2-4} \cline{6-7}
    Cox Ratio & \textbf{0.1118} &0.1091 & 0.1090 && p=0.2288  &p=0.2222 \\
    Harrell's C & \textbf{0.7873} & 0.7866 & 0.7818 && p=0.4355 & p=0.1045 \\
    Uno's C & \textbf{0.6650} & 0.6523 & 0.6441 && \textbf{p=0.0288} & \textbf{p=0.0013} \\
    Brier Point Ratio &\textbf{0.3841} & 0.3627 & 0.3516 && \textbf{p=0.0001} &\textbf{p}$\mathbf{<10^{-5} }$  \\
    Intg. Brier Ratio & \textbf{0.4451} & 0.4262 & 0.4231 && \textbf{p=0.0135} &\textbf{p=0.0055} \\
    \bottomrule
      \end{tabular}
    \caption{Average scores for OST, \pkg{rpart} and  \pkg{ctree} models on real-world datasets. The final columns show the one-sided p-values for paired t-tests comparing the outcome metrics on each instance. }
    \label{tab:realresults}
\end{table}

\begin{table}[h]
    \centering
    \begin{tabular}{l r rr}
    \toprule
    & OST & \pkg{rpart} & \pkg{ctree} \\\midrule
    Cox Ratio & 48.7 &32.8 & 36.4 \\
    Harrell's C & 57.3& 30.8&33.6\\
    Uno's C & 59.3 & 27.3& 34.1\\
    Brier Point Ratio & 56.6& 33.3 &38.4\\
    Intg. Brier Ratio & 57.6 & 30.6 &33.6 \\
    \bottomrule
      \end{tabular}
    \caption{The percentage of instances for which each algorithm was undominated by the other algorithms. Note that rows do not sum to 100, as several instances were tied. }
    \label{tab:realresultsrank}
\end{table}

 \begin{table}[h!]
 \centering
 \resizebox{\textwidth}{!}{
\begin{tabular}{@{}lcc|ccc|ccc|ccc|ccc|ccc@{}}
\toprule
 & \multicolumn{1}{l}{} & \multicolumn{1}{l}{} & \multicolumn{3}{c}{\textbf{\begin{tabular}[c]{@{}c@{}}Integrated \\ Brier Score\end{tabular}}} & \multicolumn{3}{c}{\textbf{\begin{tabular}[c]{@{}c@{}}Harrell's C\\ Score\end{tabular}}} & \multicolumn{3}{c}{\textbf{\begin{tabular}[c]{@{}c@{}}Uno's C\\ Score\end{tabular}}} & \multicolumn{3}{c}{\textbf{\begin{tabular}[c]{@{}c@{}}Cox Partial\\ Likelihood\end{tabular}}} & \multicolumn{3}{c}{\textbf{\begin{tabular}[c]{@{}c@{}}Brier Point \\ Ratio\end{tabular}}} \\ \midrule
\textbf{Dataset} & \multicolumn{1}{l}{\textbf{n}} & \multicolumn{1}{l}{\textbf{p}} & \multicolumn{1}{l}{\textbf{OST}} & \multicolumn{1}{l}{\textbf{rpart}} & \multicolumn{1}{l}{\textbf{ctree}} & \multicolumn{1}{l}{\textbf{OST}} & \multicolumn{1}{l}{\textbf{rpart}} & \multicolumn{1}{l}{\textbf{ctree}} & \multicolumn{1}{l}{\textbf{OST}} & \multicolumn{1}{l}{\textbf{rpart}} & \multicolumn{1}{l}{\textbf{ctree}} & \multicolumn{1}{l}{\textbf{OST}} & \multicolumn{1}{l}{\textbf{rpart}} & \multicolumn{1}{l}{\textbf{ctree}} & \multicolumn{1}{l}{\textbf{OST}} & \multicolumn{1}{l}{\textbf{rpart}} & \multicolumn{1}{l}{\textbf{ctree}} \\  \hline
3D Spatial Network \citep{kaul2013building} & 100000 & 1 & \textbf{0.44} & 0.33 & 0.39 & \textbf{0.82} & 0.77 & 0.79 & \textbf{0.79} & 0.73 & 0.76 & \textbf{0.05} & \textbf{0.05} & \textbf{0.05} & \textbf{0.48} & 0.35 & 0.41 \\
Airfoil Self Noise \citep{UCI} & 1503 & 4 & \textbf{0.39} & 0.33 & 0.35 & \textbf{0.83} & 0.78 & 0.78 & \textbf{0.77} & 0.7 & 0.71 & \textbf{0.09} & \textbf{0.09} & 0.08 & \textbf{0.53} & 0.42 & 0.46 \\
\begin{tabular}[c]{@{}l@{}}Appliances Energy \\ Prediction \citep{candanedo2017data}\end{tabular} & 19735 & 25 &\textbf{ 0.19} & 0.18 & 0.18 &\textbf{ 0.74} & 0.73 & \textbf{0.74} & \textbf{0.7} & 0.69 & \textbf{0.7} & \textbf{0.03} & \textbf{0.03} & \textbf{0.03} &\textbf{ 0.14 }& 0.13 & 0.12 \\
Automobile \citep{candanedo2017data} & 164 & 23 & 0.03 & \textbf{0.07} & 0.06 & 0.53 & \textbf{0.65} & 0.61 & 0.08 & \textbf{0.41} & 0.27 & 0.01 & \textbf{0.05} & 0.03 & 0 & \textbf{0.11} & \textbf{0.11} \\
Auto MPG \citep{UCI} & 398 & 7 & 0.55 & \textbf{0.56} & 0.55 & 0.85 & \textbf{0.87} &\textbf{ 0.87} & \textbf{0.79} & 0.78 & 0.77 & 0.19 & 0.2 & \textbf{0.21} & 0.58 & \textbf{0.6} & 0.58 \\
\begin{tabular}[c]{@{}l@{}}Behavior Urban \\ Traffic\end{tabular}\textbf{} & 135 & 16 & 0.18 & \textbf{0.2} & 0.18 & 0.66 & \textbf{0.67 }& 0.64 & 0.37 & \textbf{0.41} & 0.33 & 0.08 & \textbf{0.09} & 0.08 & 0.13 & \textbf{0.16} & 0.14 \\
Bike Sharing\textbf{} & 17379 & 13 & 0.92 & 0.88 & \textbf{0.93} & \textbf{0.98} & 0.96 & \textbf{0.98} & \textbf{0.96} & 0.93 &\textbf{ 0.96} & 0.15 & 0.09 & \textbf{0.2} & 0.94 & 0.91 & \textbf{0.95} \\
Blog Feedback \citep{fanaee2014event} & 52397 & 279 & \textbf{0.39 }& \textbf{0.39} & 0.38 & 0.84 & \textbf{0.85} & \textbf{0.85} & 0.79 & 0.8 & \textbf{0.82} & \textbf{0.03} & \textbf{0.03} & \textbf{0.03 }& 0.17 & \textbf{0.18} & 0.17 \\
Buzz in Social Media \citep{kawala:hal-00881395} & 100000 & 76 & \textbf{0.77} & 0.75 & \textbf{0.77} & \textbf{0.92} & 0.91 & \textbf{0.92} & \textbf{0.91} & 0.88 & 0.9 & \textbf{0.13} & 0.12 & 0.12 & \textbf{0.76 }& 0.74 & 0.75 \\
Cargo2000 \citep{6882809} & 3943 & 95 & \textbf{1} & \textbf{1} & 0.84 & \textbf{1} & \textbf{1} & 0.95 & \textbf{1} & \textbf{1} & 0.9 & \textbf{0.21} & \textbf{0.21} & 0.16 & 0.22 & \textbf{0.23} & 0.17 \\
Communities  Crime \citep{redmond2002data} & 2215 & 145 & 0.64 & 0.65 & \textbf{0.69} & 0.89 & 0.89 & \textbf{0.91} & 0.81 & 0.83 & \textbf{0.85 }& 0.17 & 0.17 & \textbf{0.19} & 0.68 & 0.7 & \textbf{0.75} \\
Computer Hardware  \citep{UCI} & 209 & 8 & \textbf{0.69} & 0.61 & 0.65 & \textbf{0.86} & 0.83 & 0.85 & \textbf{0.74} & 0.67 & 0.7 & 0.24 & 0.27 & \textbf{0.29} &\textbf{ 0.73} & 0.68 & 0.62 \\
Concrete Slump \citep{yeh2007modeling} & 103 & 6 & 0.07 & \textbf{0.14} & 0.03 & 0.62 & \textbf{0.66} & 0.56 & 0.27 & \textbf{0.35} & 0.14 & 0.04 & \textbf{0.07} & 0.02 & 0.11 & \textbf{0.13} & 0.05 \\
Concrete Strength \citep{yeh1998modeling} & 1030 & 7 & \textbf{0.42} & 0.41 & 0.4 & \textbf{0.84} & 0.83 & 0.82 & 0.74 & 0.74 & \textbf{0.75} & 0.11 &\textbf{ 0.13} & 0.12 & 0.5 & 0.47 & \textbf{0.51} \\
CSM \citep{7463737} & 232 & 11 & 0.25 & \textbf{0.32} & 0.25 & 0.71 &\textbf{ 0.76} & 0.73 & 0.48 & 0.56 & \textbf{0.57} & 0.08 & \textbf{0.11} & 0.09 & 0.34 & \textbf{0.42} & 0.26 \\
Cycle Power\textbf{} & 9568 & 3 & \textbf{0.73} & 0.71 & \textbf{0.73} & \textbf{0.92} & 0.91 & \textbf{0.92 }& \textbf{0.89} & 0.86 & \textbf{0.89} & 0.16 & 0.17 & \textbf{0.18} &\textbf{ 0.75 }& 0.72 & \textbf{0.75} \\
Electrical Stability  \citep{UCI} & 10000 & 11 & \textbf{0.4} & 0.34 & 0.39 & \textbf{0.82} & 0.79 & \textbf{0.82} & \textbf{0.79} & 0.75 & \textbf{0.79} & \textbf{0.08} & 0.06 & \textbf{0.08 }&\textbf{ 0.44} & 0.37 & \textbf{0.44} \\
Energy efficiency 1 \citep{TSANAS2012560} & 1296 & 7 & \textbf{0.95} & 0.9 & 0.9 & \textbf{0.99} & 0.97 & 0.98 & \textbf{0.98} & 0.95 & 0.93 & \textbf{0.35} & 0.3 & 0.31 & -0.11 & \textbf{-0.04} & -0.14 \\
Energy efficiency 2 \citep{TSANAS2012560} & 1296 & 7 & \textbf{0.94} & 0.9 & 0.9 & \textbf{0.99} & 0.97 & 0.97 & \textbf{0.97} & 0.95 & 0.96 & \textbf{0.27} & 0.13 & 0.21 & -0.14 & \textbf{-0.01} & -0.16 \\
Faceboook Comments \citep{Sing1503:Comment} & 40949 & 52 & \textbf{0.56} & \textbf{0.56} & 0.55 & 0.88 & 0.88 & \textbf{0.89} & 0.84 & 0.84 & \textbf{0.86} & \textbf{0.06} & \textbf{0.06} & \textbf{0.06} & -0.1 & \textbf{-0.09} & -0.11 \\
Faceboook Metrics \citep{MORO20163341} & 500 & 6 & \textbf{0.03} & 0.02 & 0.02 & 0.55 & \textbf{0.56} & 0.53 & 0.1 & \textbf{0.14} & 0.05 & \textbf{0.01} & \textbf{0.01} & \textbf{0.01} & \textbf{0.05 }& \textbf{0.05} & 0.02 \\
Fires\textbf{} & 517 & 11 & \textbf{0} & \textbf{0} & \textbf{0} & \textbf{0.5} & \textbf{0.5} &\textbf{ 0.5} &\textbf{ 0} & \textbf{0} & \textbf{0} & \textbf{0} & \textbf{0} & \textbf{0} & \textbf{0.11 }& \textbf{0.11} & \textbf{0.11} \\
GeoMusic \citep{7023456} & 1059 & 115 & 0.03 & \textbf{0.06} & 0.03 & 0.58 & \textbf{0.61} & 0.59 & 0.32 & 0.37 & \textbf{0.38} & 0.01 & \textbf{0.02} & 0.01 & 0.02 & \textbf{0.06} & 0.03 \\
\begin{tabular}[c]{@{}l@{}}Insurance Company \\ Benchmark\end{tabular} \citep{insurance} & 5822 & 84 & 0.02 & 0.02 & \textbf{0.03} & 0.59 & 0.6 & \textbf{0.62} & 0.24 & 0.25 & \textbf{0.27} & \textbf{0} & \textbf{0} & \textbf{0} & \textbf{0.33} & \textbf{0.33} & \textbf{0.33} \\
KEGG Directed \citep{UCI} & 53413 & 22 & \textbf{0.81} & 0.78 & 0.79 & \textbf{0.96} & 0.95 & 0.95 & \textbf{0.94} & 0.93 & 0.93 &\textbf{ 0.11} & \textbf{0.11} & \textbf{0.11} & 0.06 & \textbf{0.07} & \textbf{0.07} \\
KEGG Undirected \citep{UCI} & 65554 & 25 & \textbf{0.87} & 0.81 & 0.86 & 0.96 & 0.95 & \textbf{0.97} & \textbf{0.97} & 0.94 & 0.96 & 0.14 & 0.16 & \textbf{0.17 }& \textbf{0.87} & 0.81 & 0.85 \\
Kernel Performance \citep{kernel} & 100000 & 13 & \textbf{0.73} & 0.67 & 0.69 & \textbf{0.84} & 0.8 & 0.82 & \textbf{0.84} & 0.79 & 0.81 & \textbf{0.09} & 0.07 & 0.08 & \textbf{0.53} & 0.43 & 0.47 \\
Las Vegas Strip \citep{MORO201741} & 504 & 18 & \textbf{0.02} & 0 & 0.01 &\textbf{ 0.54} & 0.51 & 0.52 & \textbf{0.12} & 0.05 & 0.05 & \textbf{0} & \textbf{0} & \textbf{0} & \textbf{0.02} & -0.01 & 0 \\
Online News Popularity\textbf{} & 39644 & 58 &\textbf{ 0.05} & \textbf{0.05} & \textbf{0.05} & 0.62 & 0.62 & \textbf{0.63} & 0.56 & 0.57 & \textbf{0.58} & \textbf{0.01} & \textbf{0.01} & \textbf{0.01} & 0.16 & 0.16 & \textbf{0.17} \\
\begin{tabular}[c]{@{}l@{}}Online Video \\ Characteristics\end{tabular} \citep{UCI} & 68784 & 19 & 0.75 & 0.7 & \textbf{0.76} & \textbf{0.92} & 0.91 & \textbf{0.92} & \textbf{0.91} & 0.89 & \textbf{0.91} & 0.06 & \textbf{0.15} & \textbf{0.15 }& \textbf{0.76} & 0.73 & \textbf{0.76} \\
\begin{tabular}[c]{@{}l@{}}Optical Interconnection \\ Network\end{tabular}  \citep{Acl2015} & 640 & 8 & -0.08 & \textbf{0.32} & 0.27 & \textbf{0.81} & 0.79 & \textbf{0.81} & 0.72 & 0.67 & \textbf{0.73} & \textbf{0.12} & \textbf{0.12} & 0.11 &\textbf{ 0.7} & 0.68 & 0.69 \\
\begin{tabular}[c]{@{}l@{}}Parkinson \\ Telemonitoring\end{tabular} \citep{5339170} & 5875 & 18 & \textbf{0.59} & 0.49 & 0.42 & \textbf{0.85} & 0.82 & 0.8 & \textbf{0.8} & 0.77 & 0.73 & \textbf{0.14 }& 0.1 & 0.08 & \textbf{0.67} & 0.55 & 0.48 \\
PM2.5-Beijing \citep{doi:10.1002/2016JD024877} & 50387 & 12 &\textbf{ 0.35} & 0.32 & 0.34 & \textbf{0.8} & 0.79 & \textbf{0.8} & \textbf{0.77} & 0.75 & 0.76 & 0.05 & 0.05 & \textbf{0.06 }&\textbf{ 0.43 }& 0.4 & 0.42 \\
Propulsion Plant \citep{doi:10.1177/1475090214540874} & 11934 & 15 & \textbf{0.64 }& 0.43 & 0.28 & \textbf{0.88} & 0.8 & 0.72 & \textbf{0.87} & 0.74 & 0.57 & \textbf{0.11} & 0.08 & 0.04 & \textbf{0.7 }& 0.46 & 0.32 \\
Protein \citep{UCI} & 45730 & 8 & \textbf{0.3} & 0.26 & 0.26 &\textbf{ 0.75} & 0.73 & 0.72 & \textbf{0.72} & 0.69 & 0.69 & \textbf{0.04} & 0.03 & 0.03 & \textbf{0.32} & 0.27 & 0.27 \\
Real Estate 1 \citep{RealEstate} & 414 & 5 & \textbf{0.44} & 0.4 & 0.42 & \textbf{0.83} & 0.79 & 0.8 & \textbf{0.67} & 0.58 & 0.64 & \textbf{0.18} & 0.15 & 0.15 & \textbf{0.59} & 0.56 & 0.56 \\
Real Estate 2 \citep{RealEstate} & 53500 & 383 & \textbf{0.8} & 0.75 & 0.76 & 0.55 &\textbf{ 0.92} & 0.88 & \textbf{0.9} & 0.87 & 0.88 & 0.02 & \textbf{0.1} & 0.05 & \textbf{0.83} & 0.76 & 0.78 \\
Residential Building \citep{doi:10.1061/(ASCE)CO.1943-7862.0001047} & 372 & 107 & 0.63 & \textbf{0.64} & 0.62 & 0.86 & \textbf{0.87 }& \textbf{0.87} & 0.73 & \textbf{0.76 }& 0.74 & 0.24 & \textbf{0.27} & 0.24 & 0.63 & \textbf{0.68 }& \textbf{0.68} \\
Servo \citep{UCI} & 167 & 3 & \textbf{0.51} & 0.4 & 0.39 & \textbf{0.85} & 0.73 & 0.69 & \textbf{0.75} & 0.5 & 0.41 & \textbf{0.26} & 0.16 & 0.14 &\textbf{ 0.48} & 0.29 & 0.24 \\
Stock Market Istanbul \citep{Istanbul} & 536 & 6 & 0.11 & 0.12 & \textbf{0.14} & 0.68 & \textbf{0.69} & \textbf{0.69 }& 0.47 & \textbf{0.5} & \textbf{0.5} & 0.04 & \textbf{0.05} & \textbf{0.05} & \textbf{0.18} & 0.17 & 0.2 \\
Stock Portfolio \citep{stock} & 63 & 11 &\textbf{ 0.42} & 0.26 & 0.29 & \textbf{0.78} & 0.73 & 0.74 &\textbf{ 0.53} & 0.46 & 0.43 & \textbf{0.34} & 0.26 & 0.27 & \textbf{0.49} & 0.4 & 0.35 \\
Student Performance \citep{AHMED2016137} & 395 & 29 & 0.05 & \textbf{0.08} & \textbf{0.08 }& 0.58 & \textbf{0.61} & \textbf{0.61} & 0.2 & 0.21 & \textbf{0.26} & \textbf{0.02} & \textbf{0.02} & \textbf{0.02} & 0.07 & \textbf{0.11} & 0.1 \\
Wine Quality \citep{Cortez2009ModelingWP} & 6497 & 10 & 0.16 & 0.16 & \textbf{0.18} & 0.73 & 0.74 & \textbf{0.76} & 0.63 & 0.65 & \textbf{0.71} & 0.02 & 0.02 &\textbf{ 0.03} & \textbf{-0.04} & \textbf{-0.04} & -0.07 \\
Yacht \citep{UCI} & 308 & 5 & \textbf{0.84} & 0.8 & 0.82 &\textbf{ 0.94} & 0.91 & 0.9 & \textbf{0.85} & 0.81 & 0.77 & 0.37 & \textbf{0.41} & 0.4 & 0.8 & 0.76 & \textbf{0.82 }\\ 
\bottomrule
\end{tabular}}
\caption{Average scores for OST, \pkg{rpart}, \pkg{ctree} for each dataset across all levels of censoring.}
 \label{tab:DataSpecific}
\end{table}
 
\section{An Application to Heart Disease}\label{sec:fhs}

In this section, we provide an example of a practical application of the OST algorithm to a real-world dataset from the Framingham Heart Study. Analysis of the FHS successfully identified the common factors or characteristics that contribute to CHD using the Cox regression model \citep{cox1972regression}. In our survival tree model, we include all participants in the study  from the original cohort (1948-2014) and the offspring cohort (1971-2014) who were diagnosed with Coronary Heart Disease (CHD). The event of interest in this model is the occurrence of a myocardial infarction or stroke. All patients were followed for a period of at least 10 years after their first diagnosis of CHD and observations are marked as censored if no event was observed while the patient was under observation.

We applied our algorithm to the primary variables that have been used in the established 10-year Hard Coronary Heart Disease (HCHD) Risk Calculator and the Cardiovascular Risk Calculator \citep{AdultTreatmentPanelIII,cvd_risk}. For each participant who was diagnosed with CHD, we include the following covariates in our training dataset: gender, smoking status (smoke), Systolic Blood Pressure (SBP), Diastolic Blood Pressure (DBP), use of anti-hypertensive medication (AHT), Body Mass Index (BMI), diabetic status (diabetes). 
We did not include cholesterol levels in our analysis because these variables are highly correlated with the use of lipid lowering treatment and a high proportion of the sample population did not have sufficient data to account for this interaction.

\begin{figure}[h]
  \includegraphics[width=\textwidth]{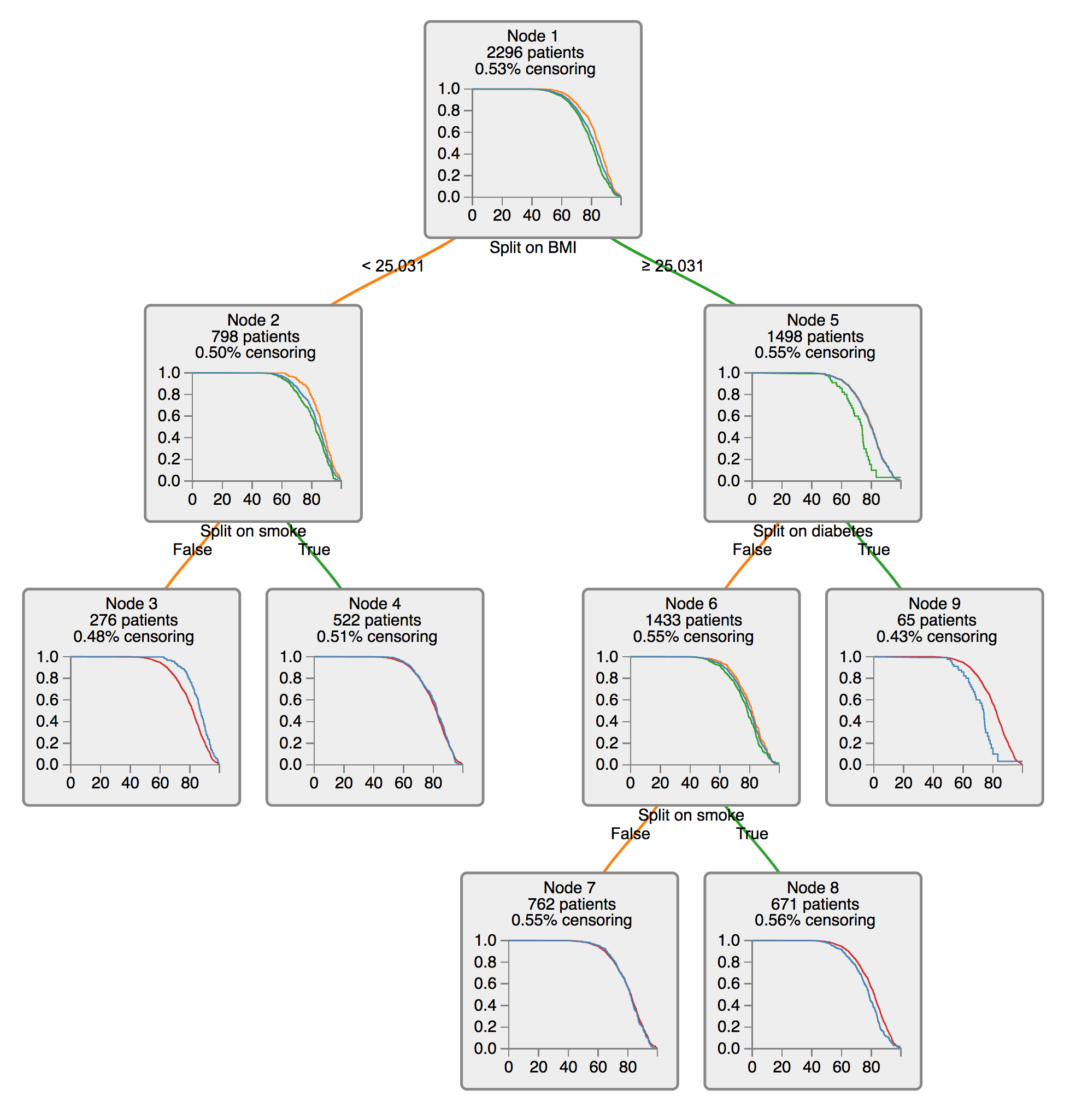}
  \caption{An illustration of Optimal Survival Trees for chd patients in the FHS.}
  \label{fig:FHS}
\end{figure}

In Figure~\ref{fig:FHS} we illustrate the output of our algorithm on the FHS dataset. Every node of the tree provides the following information:
\begin{itemize}
\item The node number.
\item Number of observations classified into the node.
\item Proportion of the node population which has been censored.
\item A plot of survival probability vs. time. In this example,  the x-axis represents age and the y-axis gives the Kaplan-Meier estimate for the probability of experiencing no adverse events.
\item Color-coded survival curves to describe the different sub-populations. In each node, the blue curves describe the individuals classified into that node. 
\item In internal (parent) nodes, the orange/green curves describe the sub-populations that are split into the left/right child node. After each split, the sub-population with higher likelihood of survival goes into the left node. 
\item In leaf nodes, the red curve shows the average survival curve for the entire tree. This facilitates easy comparisons between the survival of a specific node and the rest of the population. 
\end{itemize}

The splits illustrated in Figure~\ref{fig:FHS} include known risk factors for heart disease and are consistent with well-established medical guidelines. The algorithm identified a BMI threshold of 25 as the first split (node 1), which is in accordance with the NIH BMI ranges that classify an individual as overweight if his/her BMI is greater than or equal to 25. Multiple splits indicated a higher risk of heart attack or stroke in patients who smoke (nodes 2, 6). The group with the highest risk of an adverse event was  overweight patients with diabetes (node 5). 

\begin{figure}[h]
  \includegraphics[width=\textwidth]{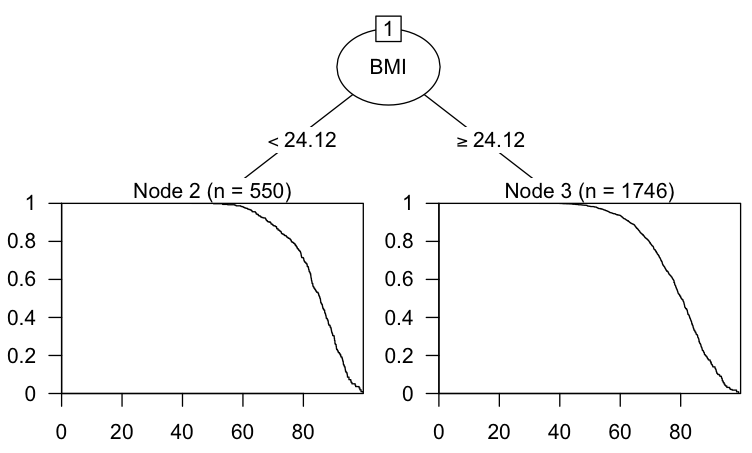}
  \caption{Illustration of the \pkg{rpart} output for chd patients in the FHS.}
  \label{fig:rpart}
\end{figure}

\begin{figure}[h]
  \includegraphics[width=\textwidth]{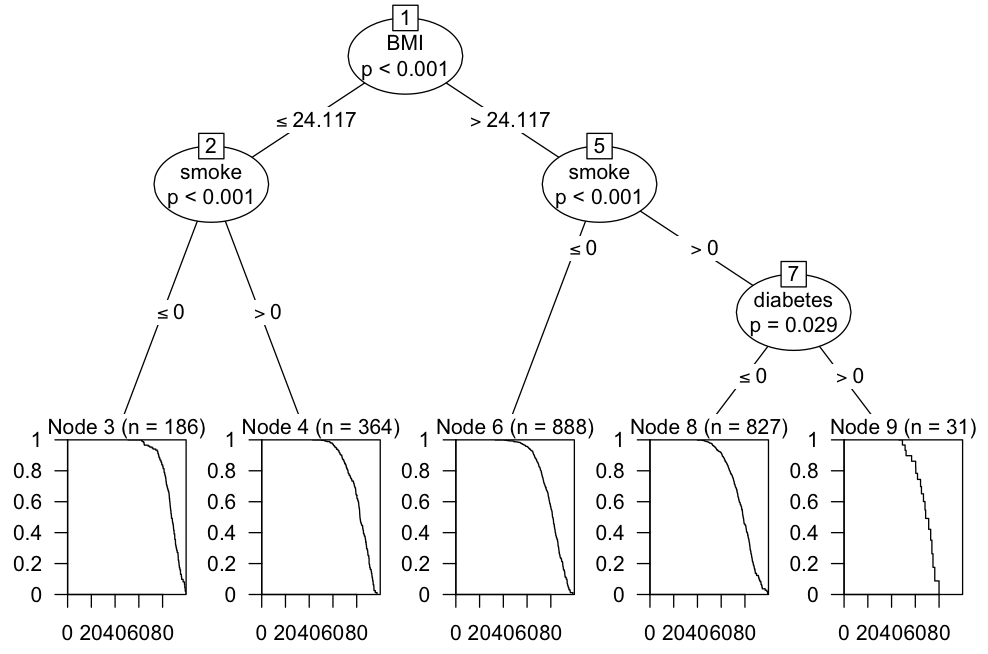}
  \caption{Illustration of the \pkg{ctree} output for chd patients in the FHS.}
  \label{fig:ctree}
\end{figure}

Figures \ref{fig:rpart} and \ref{fig:ctree} illustrate the output of the \pkg{ctree} and \pkg{rpart} algorithms applied to the same FHS population.  The \pkg{rpart} model has a single split (BMI), while the  \pkg{ctree} model contains the same variables as the OST output. The Brier scores for each model are 0.0486 (OST), 0.0249 (\pkg{rpart}) and   0.0467 (\pkg{ctree}).

The discrepancy in the Brier scores for the OST and \pkg{ctree} models is due to slight differences in the threshold and position of certain splits. For example, both methods identify that BMI is the most appropriate variable for the first split, but the BMI threshold differs. The \pkg{ctree} model sets the splitting threshold to 24.117, which is the locally optimal value for the split when building the tree greedily (the same threshold is used in the \pkg{rpart} model). By contrast, the OST algorithm selects a threshold of 25.031. This example demonstrates how  the OST algorithm's efforts to find a globally optimal solution differ from the results of locally optimal splits.  

A second difference between the tree models is the order of the smoking and diabetes splits within the overweight population. The \pkg{ctree} model splits on smoking first, since this split has the most significant p-value of the variables at node 5 in the \pkg{ctree} tree. The algorithm also recognizes that diabetes is a risk factor and incorporates this in the subsequent split. Since greedy approaches like \pkg{ctree} do not reevaluate the spits once they have been decided, the algorithm does not recognize that the overall quality of the tree can be improved by reversing the order of these splits. This discrepancy in two otherwise similar trees highlights the advantages of the more sophisticated optimization conducted by OST.

\section{Conclusion}\label{sec:conclusion}
In this paper, we have extended the state-of-the-art Optimal Trees framework to generate  interpretable models for censored data. We have also introduced a new accuracy metric, the Kaplan-Meier Area Ratio, which provides an effective way to measure the predictive power of survival tree models in simulations. 

The Optimal Survival Trees algorithm improves on the performance of existing  algorithms in terms of both classification and predictive accuracy. Our results in simulations indicate that the OST models improve consistently with increasing sample size, whereas existing algorithms are prone to overfitting in larger datasets. This is particularly important, given that the volume of medical data available for research is likely to increase significantly over the coming years.

\clearpage
\newpage
\bibliographystyle{spbasic}      
\bibliography{ref2}   

\algdef{SE}[PROCEDURE]{Procedure}{EndProcedure}%
  [2]{\ \textproc{#1}\ifthenelse{\equal{#2}{}}{}{(#2)}}%
  {}%
\newpage
\appendix 
\section{Tree simulations} \label{sec:sim_appendix}
\subsection{Tree generation algorithm}

Algorithm~\ref{alg:growtree} was used to generate ground truth models for simulated datasets.
\begin{algorithm}
\caption{Tree generation algorithm}\label{alg:growtree}
\begin{algorithmic}[1]
\State \textbf{Inputs:}
\State $\quad$ $X$  \Comment{$n \times p$ data matrix}
\State $\quad$ min\_bucket \Comment{minimum node population}
\State $\quad$ max\_depth \Comment{maximum tree depth}
\State
\Procedure{Initialize:}{}
\State $T \gets$ \{1\} \Comment{list of tree nodes, node 1 is the root node}
\State status(1) $\gets$ open \Comment{node status: open nodes may be split, closed nodes are leaf nodes}
\State population(1) $\gets$ \{1,2,\dots,n\} \Comment{observations in each node}
\State depth(1) $\gets$ 1 \Comment{depth of the node in the tree}
\EndProcedure
\Procedure{Grow Tree:}{}
\While {status($k$) = open  for any  $k \in T$} 
\State current\_node = select($k\ |\ k \in T$  and  status($k$) = open) \Comment{Select an open node to split}
\State feature\_list = permute(1:$p$+1) \Comment{Randomly order features}
\For {$j \in$ feature\_list}
\If {$j=p+1$  or  depth(current\_node) = max\_depth}
\State status(current\_node) $\gets$ closed \Comment{Close node without splitting}
\State \textbf{break} and go to (\emph{A})
\EndIf
\State feature\_values = permute(unique(\{$X_{ij}\ |\ i \in$ population(current\_node)\}))
\For {$b \in$ feature\_values} \Comment{Attempt to split on feature $j$ with threshold $b$}
\State $L_1$ = length(\{$i\ |\ i \in$ population(current\_node), $X_{ij}\leq b$ \})
\State $L_2$ = length(\{$i\ |\ i \in$ population(current\_node), $X_{ij}> b$ \})
\If {$L_1\geq$ min\_bucket \  and  \ $L_2\geq$ min\_bucket } \Comment{If split is feasible, create new nodes}
\State $T=T \bigcup$ \{total\_nodes+1,total\_nodes+2\}
\State status(total\_nodes+1) = open  
\State depth(total\_nodes+1) = depth(current\_node)+1  
\State population(total\_nodes+1) = \{$i\ |\ i \in$ population(current\_node), $X_{ij}\leq b$\} 
\State status(total\_nodes+2) = open 
\State depth(total\_nodes+2) = depth(current\_node)+1  
\State population(total\_nodes+2) = \{$i\ |\ i \in$ population(current\_node), $X_{ij}> b$\} 
\State status(current\_node) $\gets$ closed \Comment{Current node is closed}
\State \textbf{break} and go to \textproc{Grow Tree} \Comment{Select another open node to split}
\EndIf \EndFor
\EndFor \EndWhile 
\State Return $T$ \EndProcedure
\end{algorithmic}
\end{algorithm}

\subsection{Survival distributions}
Nodes in simulated trees were randomly assigned one of the following survival distributions:
\begin{itemize}
\item Exponential($\theta$):

Parameters: 0.3, 0.4, 0.6, 0.8, 0.9, 1.15, 1.5, 1.8

\item Weibull($k$,$\lambda$):

Parameters: (0.8,0.4), (0.9,0.5), (0.9,0.7), (0.9,1.1), (0.9,1.5), (1,1.1),(1,1.9),  (1.3,0.5)

\item Lognormal($\mu$,$\sigma^2$):

Parameters: (0.1,1.0), (0.2,.75), (0.3,0.3), (0.3,0.5),  (0.3,0.8), (0.4,0.32), (0.5,0.3), 
(0.5,0.7)

\item Gamma($k$,$\theta$):

Parameters: (0.2,.75), (0.3,1.3), (0.3,2), (0.5,1.5), (0.8,1.0), (0.9,1.3),  (1.4,0.9), 
(1.5,0.7)
\end{itemize}

\subsection{FHS dataset}
\noindent FHS patients inclusion criteria
\begin{itemize}
\item Participation in the Original and Offspring cohort of the FHS.
\item Formal diagnosis with chd (as indicated by the records of FHS).
\item Participants outcomes were followed for 10 consecutive years after diagnosis.
\end{itemize}

\section{Real data tests} \label{sec:real_appendix}

\subsection{Detailed results for UCI datasets}

\begin{center}
\begin{longtable}{|l|l|l|l|l|l|l|l|l|l|l|}
\caption{Dataset specific Brier Score Results for each level of censoring.} \label{tab:long1} \\

\hline \multicolumn{1}{|c|}{\textbf{Dataset}} & \multicolumn{1}{|c|}{\textbf{Method}} & \multicolumn{1}{|c|}{\textbf{0.0}}  & \multicolumn{1}{|c|}{\textbf{0.1}} & \multicolumn{1}{|c|}{\textbf{0.2}} & \multicolumn{1}{|c|}{\textbf{0.3}} & \multicolumn{1}{|c|}{\textbf{0.4}} & \multicolumn{1}{|c|}{\textbf{0.5}} & \multicolumn{1}{|c|}{\textbf{0.6}} & \multicolumn{1}{|c|}{\textbf{0.7}} & \multicolumn{1}{|c|}{\textbf{0.8}} \\ \hline 
\endfirsthead

\multicolumn{11}{c}%
{{\bfseries \tablename\ \thetable{} -- continued from previous page}} \\
\multicolumn{1}{|c|}{\textbf{Dataset}} & \multicolumn{1}{|c|}{\textbf{Method}} & \multicolumn{1}{|c|}{\textbf{0.0}}  & \multicolumn{1}{|c|}{\textbf{0.1}} & \multicolumn{1}{|c|}{\textbf{0.2}} & \multicolumn{1}{|c|}{\textbf{0.3}} & \multicolumn{1}{|c|}{\textbf{0.4}} & \multicolumn{1}{|c|}{\textbf{0.5}} & \multicolumn{1}{|c|}{\textbf{0.6}} & \multicolumn{1}{|c|}{\textbf{0.7}} & \multicolumn{1}{|c|}{\textbf{0.8}} \\ \hline 
\endhead

\hline \multicolumn{11}{|r|}{{Continued on next page}} \\ \hline
\endfoot
\hline 
\endlastfoot
                                  & OST & 0.40 & 0.42 & 0.39 & 0.43 & 0.43 & 0.47 & 0.50 & 0.48 & 0.48 \\ 
 3D Spatial Network                & rpart & 0.31 & 0.31 & 0.31 & 0.33 & 0.31 & 0.34 & 0.34 & 0.34 & 0.37 \\   
                                   & ctree & 0.36 & 0.41 & 0.39 & 0.39 & 0.39 & 0.39 & 0.41 & 0.40 & 0.35 \\   \hline
                                   & OST & 0.47 & 0.49 & 0.41 & 0.40 & 0.36 & 0.22 & 0.35 & 0.32 & 0.46 \\ 
 Airfoil  Self Noise               & rpart & 0.37 & 0.35 & 0.28 & 0.33 & 0.22 & 0.32 & 0.31 & 0.33 & 0.42 \\   
                                   & ctree & 0.39 & 0.39 & 0.37 & 0.34 & 0.32 & 0.29 & 0.26 & 0.34 & 0.46 \\   \hline
                                   & OST & 0.14 & 0.14 & 0.13 & 0.21 & 0.22 & 0.22 & 0.27 & 0.24 & 0.16 \\ 
 Appliances Energy Prediction      & rpart & 0.13 & 0.14 & 0.16 & 0.16 & 0.21 & 0.22 & 0.22 & 0.21 & 0.17 \\
\hline
\end{longtable}
\end{center}

\begin{center}
\begin{longtable}{|l|l|l|l|l|l|l|l|l|l|l|}
\caption{Dataset specific Harrell's C Score Results for each level of censoring.}
\label{tab:long2} \\

\hline \multicolumn{1}{|c|}{\textbf{Dataset}} & \multicolumn{1}{|c|}{\textbf{Method}} & \multicolumn{1}{|c|}{\textbf{0.0}}  & \multicolumn{1}{|c|}{\textbf{0.1}} & \multicolumn{1}{|c|}{\textbf{0.2}} & \multicolumn{1}{|c|}{\textbf{0.3}} & \multicolumn{1}{|c|}{\textbf{0.4}} & \multicolumn{1}{|c|}{\textbf{0.5}} & \multicolumn{1}{|c|}{\textbf{0.6}} & \multicolumn{1}{|c|}{\textbf{0.7}} & \multicolumn{1}{|c|}{\textbf{0.8}} \\ \hline 
\endfirsthead

\multicolumn{11}{c}%
{{\bfseries \tablename\ \thetable{} -- continued from previous page}} \\
\multicolumn{1}{|c|}{\textbf{Dataset}} & \multicolumn{1}{|c|}{\textbf{Method}} & \multicolumn{1}{|c|}{\textbf{0.0}}  & \multicolumn{1}{|c|}{\textbf{0.1}} & \multicolumn{1}{|c|}{\textbf{0.2}} & \multicolumn{1}{|c|}{\textbf{0.3}} & \multicolumn{1}{|c|}{\textbf{0.4}} & \multicolumn{1}{|c|}{\textbf{0.5}} & \multicolumn{1}{|c|}{\textbf{0.6}} & \multicolumn{1}{|c|}{\textbf{0.7}} & \multicolumn{1}{|c|}{\textbf{0.8}} \\ \hline 
\endhead

\hline \multicolumn{11}{|r|}{{Continued on next page}} \\ \hline
\endfoot
\hline 
\endlastfoot
                                   & OST & 0.78 & 0.79 & 0.79 & 0.80 & 0.80 & 0.83 & 0.84 & 0.85 & 0.87 \\ 
 3D Spatial Network                & rpart & 0.75 & 0.75 & 0.76 & 0.76 & 0.75 & 0.77 & 0.78 & 0.79 & 0.82 \\   
                                   & ctree & 0.75 & 0.78 & 0.78 & 0.79 & 0.78 & 0.78 & 0.80 & 0.82 & 0.81 \\   \hline
                                   & OST & 0.81 & 0.82 & 0.82 & 0.81 & 0.84 & 0.80 & 0.83 & 0.81 & 0.90 \\ 
 Airfoil  Self Noise               & rpart & 0.76 & 0.76 & 0.73 & 0.78 & 0.74 & 0.80 & 0.80 & 0.81 & 0.87 \\   
                                   & ctree & 0.72 & 0.79 & 0.79 & 0.77 & 0.78 & 0.80 & 0.78 & 0.81 & 0.80 \\   \hline
                                   & OST & 0.73 & 0.73 & 0.72 & 0.74 & 0.74 & 0.75 & 0.76 & 0.78 & 0.74 \\ 
 Appliances Energy Prediction      & rpart & 0.72 & 0.73 & 0.72 & 0.71 & 0.73 & 0.74 & 0.74 & 0.75 & 0.77 \\
\hline
\end{longtable}
\end{center}

\begin{center}
\begin{longtable}{|l|l|l|l|l|l|l|l|l|l|l|}
\caption{Dataset specific Uno's C Score Results for each level of censoring.} \label{tab:long3} \\

\hline \multicolumn{1}{|c|}{\textbf{Dataset}} & \multicolumn{1}{|c|}{\textbf{Method}} & \multicolumn{1}{|c|}{\textbf{0.0}}  & \multicolumn{1}{|c|}{\textbf{0.1}} & \multicolumn{1}{|c|}{\textbf{0.2}} & \multicolumn{1}{|c|}{\textbf{0.3}} & \multicolumn{1}{|c|}{\textbf{0.4}} & \multicolumn{1}{|c|}{\textbf{0.5}} & \multicolumn{1}{|c|}{\textbf{0.6}} & \multicolumn{1}{|c|}{\textbf{0.7}} & \multicolumn{1}{|c|}{\textbf{0.8}} \\ \hline 
\endfirsthead

\multicolumn{11}{c}%
{{\bfseries \tablename\ \thetable{} -- continued from previous page}} \\
\multicolumn{1}{|c|}{\textbf{Dataset}} & \multicolumn{1}{|c|}{\textbf{Method}} & \multicolumn{1}{|c|}{\textbf{0.0}}  & \multicolumn{1}{|c|}{\textbf{0.1}} & \multicolumn{1}{|c|}{\textbf{0.2}} & \multicolumn{1}{|c|}{\textbf{0.3}} & \multicolumn{1}{|c|}{\textbf{0.4}} & \multicolumn{1}{|c|}{\textbf{0.5}} & \multicolumn{1}{|c|}{\textbf{0.6}} & \multicolumn{1}{|c|}{\textbf{0.7}} & \multicolumn{1}{|c|}{\textbf{0.8}} \\ \hline 
\endhead

\hline \multicolumn{11}{|r|}{{Continued on next page}} \\ \hline
\endfoot

\hline \hline
\endlastfoot
                                   & OST & 0.77 & 0.78 & 0.76 & 0.77 & 0.77 & 0.80 & 0.81 & 0.82 & 0.84 \\ 
 3D Spatial Network                & rpart & 0.72 & 0.72 & 0.72 & 0.72 & 0.70 & 0.72 & 0.73 & 0.74 & 0.78 \\    
                                   & ctree & 0.73 & 0.77 & 0.76 & 0.76 & 0.75 & 0.74 & 0.77 & 0.78 & 0.76 \\    \hline
                                   & OST & 0.81 & 0.82 & 0.80 & 0.80 & 0.78 & 0.77 & 0.72 & 0.66 & 0.79 \\ 
 Airfoil  Self Noise               & rpart & 0.71 & 0.70 & 0.67 & 0.71 & 0.66 & 0.70 & 0.70 & 0.65 & 0.78 \\    
                                   & ctree & 0.71 & 0.75 & 0.74 & 0.71 & 0.67 & 0.69 & 0.66 & 0.68 & 0.81 \\    \hline
                                   & OST & 0.71 & 0.70 & 0.69 & 0.71 & 0.71 & 0.70 & 0.72 & 0.74 & 0.65 \\ 
 Appliances Energy Prediction      & rpart & 0.70 & 0.70 & 0.69 & 0.64 & 0.69 & 0.69 & 0.69 & 0.70 & 0.70 \\
\hline
\end{longtable}
\end{center}

\begin{center}
\begin{longtable}{|l|l|l|l|l|l|l|l|l|l|l|}
\caption{Dataset specific Cox Score Results for each level of censoring.}
\label{tab:long4} \\

\hline \multicolumn{1}{|c|}{\textbf{Dataset}} & \multicolumn{1}{|c|}{\textbf{Method}} & \multicolumn{1}{|c|}{\textbf{0.0}}  & \multicolumn{1}{|c|}{\textbf{0.1}} & \multicolumn{1}{|c|}{\textbf{0.2}} & \multicolumn{1}{|c|}{\textbf{0.3}} & \multicolumn{1}{|c|}{\textbf{0.4}} & \multicolumn{1}{|c|}{\textbf{0.5}} & \multicolumn{1}{|c|}{\textbf{0.6}} & \multicolumn{1}{|c|}{\textbf{0.7}} & \multicolumn{1}{|c|}{\textbf{0.8}} \\ \hline 
\endfirsthead

\multicolumn{11}{c}%
{{\bfseries \tablename\ \thetable{} -- continued from previous page}} \\
\multicolumn{1}{|c|}{\textbf{Dataset}} & \multicolumn{1}{|c|}{\textbf{Method}} & \multicolumn{1}{|c|}{\textbf{0.0}}  & \multicolumn{1}{|c|}{\textbf{0.1}} & \multicolumn{1}{|c|}{\textbf{0.2}} & \multicolumn{1}{|c|}{\textbf{0.3}} & \multicolumn{1}{|c|}{\textbf{0.4}} & \multicolumn{1}{|c|}{\textbf{0.5}} & \multicolumn{1}{|c|}{\textbf{0.6}} & \multicolumn{1}{|c|}{\textbf{0.7}} & \multicolumn{1}{|c|}{\textbf{0.8}} \\ \hline 
\endhead

\hline \multicolumn{11}{|r|}{{Continued on next page}} \\ \hline
\endfoot
\hline 
\endlastfoot
                                   & OST & 0.04 & 0.04 & 0.03 & 0.06 & 0.02 & 0.06 & 0.08 & 0.06 & 0.09 \\ 
 3D Spatial Network                & rpart & 0.03 & 0.04 & 0.04 & 0.04 & 0.04 & 0.05 & 0.05 & 0.06 & 0.08 \\    
                                   & ctree & 0.04 & 0.05 & 0.05 & 0.05 & 0.05 & 0.05 & 0.06 & 0.07 & 0.07 \\    \hline
                                   & OST & 0.05 & 0.05 & 0.07 & 0.05 & 0.14 & 0.06 & 0.13 & 0.12 & 0.11 \\ 
 Airfoil  Self Noise               & rpart & 0.07 & 0.07 & 0.06 & 0.08 & 0.06 & 0.10 & 0.10 & 0.11 & 0.18 \\    
                                   & ctree & 0.05 & 0.09 & 0.09 & 0.09 & 0.09 & 0.10 & 0.08 & 0.11 & 0.03 \\    \hline
                                   & OST & 0.02 & 0.02 & 0.02 & 0.03 & 0.03 & 0.03 & 0.04 & 0.05 & 0.04 \\ 
 Appliances Energy Prediction      & rpart & 0.02 & 0.02 & 0.02 & 0.02 & 0.03 & 0.03 & 0.04 & 0.04 & 0.05 \\
\hline
\end{longtable}
\end{center}

\begin{center}
\begin{longtable}{|l|l|l|l|l|l|l|l|l|l|l|}
\caption{Dataset specific Brier Point Ratio Results for each level of censoring.}
\label{tab:long5} \\

\hline \multicolumn{1}{|c|}{\textbf{Dataset}} & \multicolumn{1}{|c|}{\textbf{Method}} & \multicolumn{1}{|c|}{\textbf{0.0}}  & \multicolumn{1}{|c|}{\textbf{0.1}} & \multicolumn{1}{|c|}{\textbf{0.2}} & \multicolumn{1}{|c|}{\textbf{0.3}} & \multicolumn{1}{|c|}{\textbf{0.4}} & \multicolumn{1}{|c|}{\textbf{0.5}} & \multicolumn{1}{|c|}{\textbf{0.6}} & \multicolumn{1}{|c|}{\textbf{0.7}} & \multicolumn{1}{|c|}{\textbf{0.8}} \\ \hline 
\endfirsthead

\multicolumn{11}{c}%
{{\bfseries \tablename\ \thetable{} -- continued from previous page}} \\
\multicolumn{1}{|c|}{\textbf{Dataset}} & \multicolumn{1}{|c|}{\textbf{Method}} & \multicolumn{1}{|c|}{\textbf{0.0}}  & \multicolumn{1}{|c|}{\textbf{0.1}} & \multicolumn{1}{|c|}{\textbf{0.2}} & \multicolumn{1}{|c|}{\textbf{0.3}} & \multicolumn{1}{|c|}{\textbf{0.4}} & \multicolumn{1}{|c|}{\textbf{0.5}} & \multicolumn{1}{|c|}{\textbf{0.6}} & \multicolumn{1}{|c|}{\textbf{0.7}} & \multicolumn{1}{|c|}{\textbf{0.8}} \\ \hline 
\endhead

\hline \multicolumn{11}{|r|}{{Continued on next page}} \\ \hline
\endfoot
\hline 
\endlastfoot
                                   & OST & 0.44 & 0.46 & 0.45 & 0.47 & 0.48 & 0.53 & 0.51 & 0.49 & 0.45 \\ 
 3D Spatial Network                & rpart & 0.36 & 0.36 & 0.37 & 0.39 & 0.33 & 0.35 & 0.34 & 0.35 & 0.33 \\    
                                   & ctree & 0.38 & 0.46 & 0.43 & 0.46 & 0.42 & 0.39 & 0.41 & 0.41 & 0.33 \\    \hline
                                   & OST & 0.52 & 0.52 & 0.54 & 0.61 & 0.58 & 0.53 & 0.48 & 0.46 & 0.52 \\ 
 Airfoil  Self Noise               & rpart & 0.37 & 0.41 & 0.38 & 0.52 & 0.33 & 0.47 & 0.47 & 0.44 & 0.41 \\    
                                   & ctree & 0.41 & 0.46 & 0.52 & 0.49 & 0.45 & 0.47 & 0.44 & 0.42 & 0.51 \\    \hline
                                   & OST & 0.20 & 0.21 & 0.16 & 0.18 & 0.23 & 0.05 & 0.06 & 0.22 & -0.06 \\ 
 Appliances Energy Prediction      & rpart & 0.18 & 0.20 & 0.18 & 0.18 & 0.24 & 0.02 & 0.03 & 0.19 & -0.04 \\
\hline
\end{longtable}
\end{center}

\subsection{Aggregate Results for different Censoring Levels}

\begin{table}[h]
\centering
 \resizebox{\textwidth}{!}{
%
}} \\ \midrule
\textbf{\% Censoring} & \textbf{Method} & \textbf{Mean} & \textbf{Std. Dev.} & \textbf{Mean} & \textbf{Std. Dev.} & \textbf{Mean} & \textbf{Std. Dev.} & \textbf{Mean} & \textbf{Std. Dev.} & \textbf{Mean} & \textbf{Std. Dev.} \\
 & ctree & 0.43 & 0.047 & 0.769 & 0.022 & 0.665 & 0.041 & 0.097 & 0.016 & 0.325 & 0.068 \\
0 & OST & \textbf{0.48} & 0.049 & \textbf{0.79} & 0.023 & \textbf{0.704} & 0.043 & \textbf{0.1} & 0.018 & \textbf{0.366} & 0.07 \\
 & rpart & 0.404 & 0.042 & 0.767 & 0.018 & 0.654 & 0.035 & 0.084 & 0.011 & 0.332 & 0.055 \\ \hline
 & ctree & 0.422 & 0.046 & 0.776 & 0.021 & 0.656 & 0.042 & \textbf{0.096} & 0.015 & 0.323 & 0.067 \\
0.1 & OST & \textbf{0.464} & 0.046 & \textbf{0.788} & 0.022 & \textbf{0.687} & 0.041 & 0.087 & 0.012 & \textbf{0.352} & 0.067 \\
 & rpart & 0.403 & 0.043 & 0.767 & 0.02 & 0.636 & 0.04 & 0.075 & 0.01 & 0.323 & 0.053 \\\hline
 & ctree & 0.423 & 0.046 & 0.776 & 0.021 & 0.647 & 0.043 & 0.091 & 0.014 & 0.313 & 0.067 \\
0.2 & OST & \textbf{0.443} & 0.049 & 0.778 & 0.022 & \textbf{0.669} & 0.041 & \textbf{0.097} & 0.014 & \textbf{0.349} & 0.068 \\
 & rpart & 0.421 & 0.045 & \textbf{0.781} & 0.019 & 0.66 & 0.038 & 0.093 & 0.013 & 0.339 & 0.066 \\\hline
 & ctree & 0.43 & 0.045 & 0.779 & 0.021 & 0.644 & 0.042 & 0.097 & 0.013 & 0.316 & 0.064 \\
0.3 & OST & 0.399 & 0.074 & 0.772 & 0.023 & 0.642 & 0.045 & \textbf{0.1} & 0.015 & \textbf{0.331} & 0.065 \\
 & rpart & \textbf{0.434} & 0.045 & \textbf{0.784} & 0.02 & \textbf{0.659} & 0.038 & 0.094 & 0.014 & 0.315 & 0.063 \\\hline
 & ctree & 0.434 & 0.046 & 0.778 & 0.022 & 0.635 & 0.044 & 0.097 & 0.013 & 0.307 & 0.065 \\
0.4 & OST & \textbf{0.442} & 0.048 & 0.774 & 0.023 & 0.639 & 0.045 & \textbf{0.105} & 0.015 & \textbf{0.346} & 0.066 \\
 & rpart & 0.429 & 0.049 & \textbf{0.784} & 0.021 & \textbf{0.646} & 0.042 & 0.104 & 0.015 & 0.332 & 0.066 \\\hline
 & ctree & 0.429 & 0.047 & 0.784 & 0.021 & 0.641 & 0.042 & 0.109 & 0.014 & 0.387 & 0.055 \\
0.5 & OST & 0.428 & 0.061 & \textbf{0.793} & 0.022 & \textbf{0.674} & 0.041 & \textbf{0.12} & 0.017 & \textbf{0.418} & 0.056 \\
 & rpart & \textbf{0.443} & 0.046 & 0.782 & 0.022 & 0.635 & 0.043 & 0.108 & 0.015 & 0.384 & 0.055 \\\hline
 & ctree & 0.422 & 0.047 & 0.787 & 0.021 & 0.637 & 0.043 & 0.114 & 0.014 & 0.434 & 0.057 \\
0.6 & OST & \textbf{0.455} & 0.049 & 0.781 & 0.024 & 0.645 & 0.048 & \textbf{0.118} & 0.016 & \textbf{0.456} & 0.058 \\
 & rpart & 0.439 & 0.047 & \textbf{0.791} & 0.022 & \textbf{0.648} & 0.043 & \textbf{0.118} & 0.015 & 0.436 & 0.057 \\\hline
 & ctree & 0.429 & 0.047 & 0.797 & 0.022 & 0.642 & 0.044 & 0.136 & 0.02 & 0.404 & 0.091 \\
0.7 & OST & \textbf{0.461} & 0.048 & 0.803 & 0.024 & 0.65 & 0.048 & 0.132 & 0.02 & \textbf{0.438} & 0.092 \\
 & rpart & 0.439 & 0.046 & \textbf{0.807} & 0.021 & \textbf{0.668} & 0.04 & \textbf{0.138} & 0.018 & 0.425 & 0.09 \\\hline
 & ctree & 0.39 & 0.049 & 0.792 & 0.023 & 0.629 & 0.046 & 0.145 & 0.023 & 0.357 & 0.106 \\
0.8 & OST & \textbf{0.435} & 0.05 & 0.806 & 0.024 & \textbf{0.675} & 0.045 & 0.146 & 0.027 & \textbf{0.402} & 0.108 \\
 & rpart & 0.424 & 0.047 & \textbf{0.816} & 0.021 & 0.667 & 0.043 & \textbf{0.168} & 0.023 & 0.378 & 0.105 \\ \bottomrule
\end{tabular}}
\caption{Average scores for OCT, \pkg{rpart}, \pkg{ctree} models for real world datasets for each level of censoring.}
\label{tab:detailedCensResults}
\end{table}

\end{document}